\documentclass{article}
\usepackage[nonatbib,preprint]{neurips_data_2021}

\usepackage[table]{xcolor}
\usepackage[utf8]{inputenc} 
\usepackage[T1]{fontenc}    
\usepackage{hyperref}       
\usepackage{url}  
\usepackage[utf8]{inputenc}
\usepackage{amsmath}
\usepackage{amssymb}
\usepackage{graphicx}
\usepackage{subcaption}
\usepackage{rotating}
\usepackage{tabulary}
\usepackage{algorithm}
\usepackage{algorithmic}
\usepackage{hyperref}
\usepackage{todonotes}
\usepackage[thmmarks, thref, amsmath]{ntheorem}
\usepackage{todonotes}
\usepackage{wrapfig}
\usepackage[toc,page,header]{appendix}
\usepackage{minitoc}
\newtheorem*{lemma-nono}{Lemma}
\newtheorem*{theorem-nono}{Theorem}
\theoremstyle{plain}

\theoremheaderfont{\itshape}
\theorembodyfont{\upshape}
\theoremstyle{plain}

\theoremstyle{nonumberplain}
\theoremheaderfont{\scshape}
\theorembodyfont{\upshape}
\usepackage{enumitem,lipsum}
\DeclareMathOperator*{\argmax}{argmax}
\theorempostwork{\setcounter{case}{0}}

\title{Effective Evaluation of Deep Active Learning on Image Classification Tasks}

\author{Nathan Beck\textsuperscript{1} \space\space\space\space Durga Sivasubramanian\textsuperscript{2} \space\space\space\space Apurva Dani\textsuperscript{3} \\
\textbf{Ganesh Ramakrishnan\textsuperscript{2} \space\space\space\space Rishabh Iyer\textsuperscript{1}} \\
\textsuperscript{1}The University of Texas at Dallas\space\space\space\space \textsuperscript{2}Indian Institute of Technology, Bombay\\ \textsuperscript{3}AIFY Innovation Labs \\
\texttt{\{nathan.beck, rishabh.iyer\}@utdallas.edu} \space\space\space\space
\texttt{\{durgas, ganesh\}@cse.iitb.ac.in}\\ \texttt{apurvadani@gmail.com}
}

\begin{document}

\maketitle
\doparttoc
\faketableofcontents
\begin{abstract}
With the goal of making deep learning more label-efficient, a growing number of papers have been studying active learning (AL) for deep models. However, there are a number of issues in the prevalent experimental settings, mainly stemming from a lack of unified implementation and benchmarking. Issues in the current literature include sometimes contradictory observations on the performance of different AL algorithms, unintended exclusion of important generalization approaches such as data augmentation and SGD for optimization, a lack of study of evaluation facets like the labeling efficiency of AL, and little or no clarity on the scenarios in which AL outperforms random sampling (RS). In this work, we present a unified re-implementation of state-of-the-art AL algorithms in the context of image classification via our new open-source AL toolkit DISTIL\footnote{\href{https://github.com/decile-team/distil}{\color{blue}https://github.com/decile-team/distil}}, and we carefully study these issues as facets of effective evaluation. On the positive side, we show that AL techniques are $2\times$ to $4\times$ more label-efficient compared to RS with the use of data augmentation. Surprisingly, when data augmentation is included, there is no longer a consistent gain in using \textsc{Badge}, a state-of-the-art approach, over simple uncertainty sampling. We then do a careful analysis of how existing approaches perform with varying amounts of redundancy and number of examples per class. Finally, we provide several insights for AL practitioners to consider in future work, such as the effect of the AL batch size, the effect of initialization, the importance of retraining the model at every round, and other insights. \looseness-1 

\end{abstract}
\section{Introduction}

Much of deep learning owes its success to the vast quantities of data used in training deep neural networks. Indeed, data sets used in training deep neural networks range from tens of thousands of data instances to millions of data instances. While using these large data sets continues to offer substantive performance benefits, the act of procuring labels for these data sets can be expensive and time-consuming depending on the task at hand. For example, medical imaging tasks for predicting cancer require the specialized skill-set of a radiologist, whose time is both hard to get and very expensive. Active learning (AL) tries to address this problem by sampling the smallest number of labeled examples needed to reach a desired test accuracy. In AL, the unlabeled data instances are intelligently chosen by a selection algorithm for a human annotator to label and present to an existing model. This selection is performed by using measures like uncertainty, diversity, or a combination.\looseness-1 


Despite the amazing progress by various papers in deep AL, a number of issues remain. Firstly, a number of contradictory findings have been recorded in past work~\cite{munjal2020robust}, including large differences in accuracies of certain baselines and contradictory conclusions on the relative performance of certain algorithms ({\em e.g.}, coreset approach~\cite{sener2018active} compared to random and uncertainty sampling). A lot of this is because different implementation details are often used ({\em e.g.}, differences in hyper-parameters, learning algorithms, {\em etc.}).
Additionally, different papers have used different sets of baselines, and a few important baselines ({\em e.g.},~\cite{pmlr-v37-wei15}) have rarely been used in these comparisons. Furthermore, most AL publications do not use data augmentation~\cite{cubuk2019autoaugment,cubuk2019randaugment,perez2017effectiveness,zagoruyko2017wide}, which significantly improves the generalization performance of deep models. The lack of data augmentation in most AL approaches such as those in~\cite{ash2020deep,sener2018active,sinha2019variational} is often paired with the use of adaptive optimizers such as Adam~\cite{kingma2017adam} and RMSProp, which have been shown to hamper generalization performance~\cite{zhou2020theoretically, wilson2017marginal} in deep learning. 
Moreover, a number of important details and guiding principles for practitioners are not clear: \textbf{a)} Should model training be restarted after every round, or can one fine-tune the current model? \textbf{b)} What is the effect of using carefully crafted seed sets in AL? \textbf{c)} Is there an ideal AL batch size, or does the batch size actually matter? \textbf{d)} When is AL more effective than random sampling, and what are the factors that affect the labeling efficiency of AL algorithms? Lastly, \textbf{e)} how scalable are AL algorithms; specifically, what is the amount of time taken by model training versus that taken by AL selection, and can this be made more compute and energy-efficient? We study all these questions in the context of image classification tasks.
%
To address the issues pointed out above, we provide a unified and modular re-implementation of most AL algorithms, building upon~\cite{DeepALRepo,badgerepo,NEURIPS2019_9015}. Below are some of the key findings and takeaways:
\begin{enumerate}[leftmargin=*,itemsep=-0.5ex]
    \item Data augmentation and other generalization approaches have a considerable impact on the test performance as well as on the labeling efficiency. For a ResNet-18~\cite{he2015deep} model applied on the CIFAR-10~\cite{Krizhevsky09learningmultiple} data set, we show that most AL approaches with data augmentation achieve $> 90\%$ accuracy with around 13k labeled points for CIFAR-10 while achieving labeling efficiencies of $2\times$ to $4\times$ on most data sets. \emph{Data augmentation not only improves the test performance significantly but also improves the labeling efficiency with respect to RS.}
    \item We observe that 
    \emph{SGD performs and generalizes better than Adam~\cite{kingma2017adam} consistently in AL} (which corroborates past work~\cite{zhou2020theoretically,wilson2017marginal}). 
    \item In the presence of data augmentation and the SGD optimizer, there is no significant advantage of diversity over very simple uncertainty sampling (at least in most standard academic data sets). \emph{Specifically, the state-of-the-art approach~\textsc{Badge}~\cite{ash2020deep} does not consistently outperform uncertainty sampling approaches such as entropy~\cite{settles.tr09} and least confidence~\cite{settles.tr09}.} 
    \item When we make the data set artificially 
    redundant (either by repeating data points or by using near repetitions through augmentations),  we see that \textsc{Badge} starts outperforming uncertainty sampling. \emph{The difference between the two increases as the amount of redundancy increases, which suggests that for data sets with a lot of redundancy ({\em e.g.}, frames from a video), it makes sense to use an approach such as \textsc{Badge}.}\looseness-1
    \item We see that the number of instances per class has an important impact on the performance of AL algorithms: \emph{The fewer the examples there are per class, the smaller the room there is for AL to improve over random sampling}, which further leads to reduced labeling efficiency.
    \item We study the role of labeled set initialization and the choice of the AL batch size. We find that the initialization of the labeled seed set ({\em e.g.}, a random versus a more diverse or representative seed set) has little to no impact on the performance of AL after a few rounds; likewise, reasonably sized choices of the AL batch size also have little to no impact. 
    \item We observe that updating the models from previous rounds (fine-tuning) versus retraining the models from scratch negatively impacts the performance of AL \emph{only in the early selection rounds} -- the performance difference between the two strategies vanishes once the model has stabilized.\looseness-1 
    \item Finally, we study the turnaround time and the energy consumption of AL. We find that the most time-consuming and energy-inefficient part of the AL loop is the model (re)training. We suggest possible strategies to accelerate this component of AL by applying our findings on the impact of AL batch size and model fine-tuning, and we explore the use of data subset selection (DSS) techniques~\cite{killamsetty2021gradmatch} for accelerating model training. We show that these improve the training time by $3\times$ with negligible accuracy loss. 
\end{enumerate}

Hence, we contribute a series of observations on our chosen facets of evaluation that can help address shortcomings in effective evaluation of active learning. Via our contribution, we hope to encourage future practitioners to evaluate AL using our observations on the evaluation facets listed in \hyperref[sec:facets]{\textbf{Sec. 2}}.\looseness-1

\section{Facets of Effective Evaluation}
\label{sec:facets}

To set the context for our key findings, we enumerate some facets of effective evaluation of AL.

\label{sec:p1}
\textbf{P1: Unified Experimental Setting.} An effective comparison of AL selection algorithms requires attention to the experimental setting. Since several experimental parameters such as the choice of optimizer, the learning rate and its schedule, AL batch sizes, labeled set initializations, and other parameters have varied across publications, it is important to present a unified experimental setting wherein many of the re-implemented AL approaches can be compared. Luckily, many AL approaches can be decoupled from the training loop since these methods only consider the batch query. In this fashion, we are able to control all aspects of intermediate training and common parameters used in AL batch selection. In this work, we utilize {\color{blue}\href{https://github.com/decile-team/distil}{DISTIL}}, our unified re-implementation of most state-of-the-art AL approaches, in a number of experimental settings. We build upon the Deep Active Learning GitHub repository~\cite{DeepALRepo,badgerepo}, add several recent algorithms~\cite{pmlr-v37-wei15,killamsetty2020glister,sinha2019variational}, and make the code modular to enable the use of different training loops. We provide the code along with Jupyter notebooks to reproduce the experiments from this paper in \hyperref[sec:a1]{Appendix A}. Detailed results are in \hyperref[sec:e1]{\textbf{Sec. 4.1}}.\looseness-1

\label{sec:p2}
\textbf{P2: Emphasis on Labeling Efficiency.} Most AL works present their results by comparing the test accuracy with respect to the number of labeled points. While this perspective highlights the benefit of a selection algorithm compared to their baselines, it does not explicitly show the labeling efficiency of that algorithm with respect to random sampling. Since AL is premised on obtaining a target test accuracy using the fewest labels possible, it must be clear how many fewer labels a selection algorithm needs in order to achieve that target test accuracy with respect to random sampling. In this paper (specifically in ~\hyperref[sec:e1]{\textbf{Sec. 4.1}} and~\hyperref[sec:e2]{\textbf{Sec. 4.2}}), \emph{we emphasize this measure through the use of labeling efficiency plots and motivate the need for studying labeling efficiency in future works as a means to more holistically understand AL selection algorithms.} We give a concise definition of labeling efficiency in~\hyperref[sec:e1]{\textbf{Sec. 4.1.}}\looseness-1 

\label{sec:p3}
\textbf{P3: AL with Data Augmentation.} Next, we study the effect of data augmentation~\cite{cubuk2019autoaugment, cubuk2019randaugment, perez2017effectiveness, zagoruyko2017wide}. Data augmentation has become commonplace in deep learning but has hardly been studied or used in AL algorithms; most AL papers do not report results with data augmentation. For example, the recent state-of-the-art \textsc{Badge}~\cite{ash2020deep} approach achieved only slightly more than 80\% towards the end of AL whereas state-of-the-art training methods (with data augmentation) achieve more than $90\%$ for most model families. We show that data augmentation approaches such as random cropping, random horizontal flips, and other transformations when used in AL training loops help achieve $90\%$ accuracy with only 13k labeled points in CIFAR-10 (26\% of the data set). \emph{Data augmentation not only increases the accuracies achieved using AL but also improves the labeling efficiency. Somewhat surprisingly, we also see that \textsc{Badge}-like diversity-based approaches no longer outperform simple entropy sampling~\cite{settles.tr09}.} We hypothesize that the augmentations help provide diversity to the selected data instances in the low-redundancy settings offered by the data sets commonly used in academia. 
We report these experiments in \hyperref[sec:e2]{\textbf{Sec. 4.2}}.\looseness-1

\label{sec:p4}
\textbf{P4: AL with SGD and Other Generalization Approaches.} Another facet of effective evaluation of AL involves analyzing the effect of the optimizer. Notably, a common observation is that Adam~\cite{kingma2017adam} generalizes poorly compared to SGD~\cite{wilson2017marginal,zhou2020theoretically}, but we also find that most AL approaches~\cite{ash2020deep,sener2018active,sinha2019variational} have used adaptive optimizers like Adam and RMSProp. Knowing this fact and that data augmentation helps the generalization performance of AL, we ask if {\it choosing SGD over Adam~\cite{kingma2017adam} also improves the generalization performance of AL.} We explore how SGD compares against Adam and find that \emph{Adam often results in reduced labeling efficiency and generalization performance despite being faster to converge than SGD} (\hyperref[sec:e3]{\textbf{Sec. 4.3}}). To complement our observations concerning the generalization benefit of SGD, we also explore the effect of stochastic weight averaging~\cite{izmailov2019averaging} and shake-shake regularization~\cite{gastaldi2017shakeshake} in AL, noting that both techniques further increase generalization performance. We report this in \hyperref[sec:a3]{Appendix C}.\looseness-1 

\label{sec:p5}
\textbf{P5: Redundancy.} As a result of the growth of deep learning data sets, redundant data has become a more prevalent phenomenon in deep learning. However, many of the experiments performed to assess the quality of AL algorithms neglect accounting for redundant data sets. As a result, the reported performance of these algorithms might not be indicative of their actual performance in real-world settings. We carefully examine the effect of redundancy in AL algorithms in \hyperref[sec:e4]{\textbf{Sec. 4.4}} by progressively duplicating points in the CIFAR-10 data set~\cite{Krizhevsky09learningmultiple}. We also consider another similar redundancy setting in \hyperref[sec:a4]{Appendix D} where we instead add augmented copies of data instances with subtle random translations and random cropping. We perform all experiments \emph{with data augmentation} in the training. \emph{We note that diversity-based methods such as \textsc{Badge}~\cite{ash2020deep} are robust against redundant data while uncertainty-based methods such as entropy sampling~\cite{settles.tr09} poorly handle redundant data.}\looseness-1

\label{sec:p6}
\textbf{P6: When does AL offer benefit compared to random sampling?} Most past works in AL have not provided guidelines as to when AL yields benefit over random sampling. In this paper, we hypothesize that the number of instances per class in the unlabeled data set provides a cue as to how much benefit AL can have. We are motivated by a simple observation (as we show in \hyperref[sec:e1]{\textbf{Sec. 4.1}}) that AL achieves a labeling efficiency of $1.8\times$ to $2\times$ on CIFAR-10~\cite{Krizhevsky09learningmultiple} but only $1.3\times$ on CIFAR-100~\cite{Krizhevsky09learningmultiple}. To provide a clear experiment to test this hypothesis, we explore a scenario in \hyperref[sec:e5]{\textbf{Sec. 4.5}} where we restrict the number of examples in CIFAR-10~\cite{Krizhevsky09learningmultiple} on a per-class basis in the unlabeled set. We study the effect that this restriction has on the labeling efficiency of AL, \emph{noting that having fewer instances per class results in reduced labeling efficiency.}

\label{sec:p7}
\textbf{P7: AL Batch Size and Seed Set Initialization.} Across AL publications, different choices for the AL batch size are used. To draw an effective evaluation of AL, the effect of the AL batch size needs to be studied. Namely, is there a difference in the accuracies obtained by AL in practical scenarios if the AL batch size is varied? Furthermore, the labeled seed sets used in many AL experiments are often random; however, other sophisticated methods to construct higher-quality labeled seed sets ({\em e.g.}, representative seed sets) can be used. In this work, we examine the effect of the AL batch size and the labeled seed set on the accuracies obtained by AL. These experiments are conducted in \hyperref[sec:e6]{\textbf{Sec. 4.6}}; the high-level takeaway is that \emph{neither reasonable choices of the AL batch size nor the initial seed set play a significant role in the performance (accuracies and labeling efficiencies) of AL}.\looseness-1

\label{sec:p8}
\textbf{P8: Model Updating versus Model Retraining.} Another facet of effective evaluation of AL involves analyzing the effect of updating the model between AL rounds versus retraining the model from a random initialization. Previous work done in~\cite{ash2020warmstarting} suggests that warm-starting in deep learning can hurt the generalization performance of these models; however, their study was performed without data augmentation and with the Adam optimizer. We explore the effect of model updating between selection rounds while using SGD and data augmentation. \emph{We note that the test performance is negatively affected for the early rounds of AL when doing model updating versus model retraining; however, this negative effect diminishes once the model has matured and stabilized.} This experiment is conducted in \hyperref[sec:e7]{\textbf{Sec. 4.7}}. Furthermore, we observe that when we use Adam and no data augmentation (\hyperref[sec:a5]{Appendix E}), we see a larger and more consistent degradation of the test performance of model updating versus model retraining, which confirms the observation of~\cite{ash2020warmstarting}. 

\label{sec:p9}
\textbf{P9: Scalability.} AL attempts to solve the labeling problem in large data sets, and large-scale usages of AL require that both the selection algorithm and the training procedure scale to the needs of these large data sets. Hence, we study the time spent in computing the query selection and in performing the intermediate model training, \emph{observing that the dominating time factor of most AL configurations is often the intermediate model training.} Accordingly, scalable AL usages can then draw upon our previous observations concerning AL batch size and model updating. Namely, utilizing larger AL batch sizes whenever possible can reduce the number of intermediate training rounds needed, and opting for model updating can reduce the amount of time spent in intermediate model training. To alleviate the possible performance degradation that accompanies model updating, we further study the use of data subset selection (DSS) techniques to accelerate training. Specifically, we utilize \textsc{Grad-Match}~\cite{killamsetty2021gradmatch} in most intermediate training rounds, occasionally performing full training in a few intermediate rounds to evaluate the performance of each selection algorithm. We show in \hyperref[sec:e8]{\textbf{Sec. 4.8}} that, on top of achieving significant speedups, \emph{there is little to no performance degradation by using some of the above approaches with AL}.\looseness-1
\section{Deep Active Learning Algorithms}\label{sec:al-algos}
Similar to supervised learning, AL starts with a labeled seed set that is used to train an initial model. Unlike supervised learning, there is another set of unlabeled instances $\mathcal{D_{UL}}$ from which AL periodically selects instances to label and add to the existing labeled set. The AL algorithm uses information from the model, from the unlabeled set, or from both when selecting these instances to yield the largest increase in test performance after the model is updated or retrained. To better understand the main approaches compared in this paper, we introduce some notation that we will use in our description of each algorithm. Let $f$ denote the deep neural network used in training, $f_l$ denote the output of $f$ at layer $l$, and $\theta_l$ denote the parameters of layer $l$. We use $b$ to denote the AL batch size and $L$ to denote the number of layers of $f$.\looseness-1

\textbf{Uncertainty sampling} selects instances from $\mathcal{D_{UL}}$ that maximize a measure of uncertainty on the model's prediction of the instance's label. Common uncertainty-based sampling techniques are \textbf{entropy}, \textbf{least confidence}, and \textbf{margin sampling} techniques~\cite{settles.tr09}. Using $p(x) = \text{softmax}(f_L(x))$, each computes the following for each $x\in \mathcal{D_{UL}}$:

\begin{center}
\begin{tabular}{l l}
    \textbf{Entropy:} & $H(x) = -\sum_i p(x)_i\log p(x)_i$\\
    \textbf{Least Confidence:} & $C(x) = \max_i p(x)_i$\\
    \textbf{Margin:} & $M(x) = p(x)_{\sigma_1} - p(x)_{\sigma_2}$
\end{tabular}
\end{center}

where $\sigma_1$ and $\sigma_2$ denote the indices of the largest and second-largest components of $p(x)$, respectively. \textbf{Entropy sampling} chooses $x\in \mathcal{D_{UL}}$ associated with the $b$ largest $H(x)$ values. \textbf{Least confidence sampling} and \textbf{margin sampling} choose $x\in \mathcal{D_{UL}}$ associated with the $b$ smallest $C(x)$ and $M(x)$ values, respectively. By selecting uncertain points, future training on $f$ can further refine the decision boundary, giving improved performance.\looseness-1 

Next, the \textbf{\textsc{CoreSet}} approach tries to find instances that represent or capture the structure of $\mathcal{D_{UL}}$ by minimizing a core-set loss~\cite{sener2018active}. The authors of~\cite{sener2018active} show that this can be achieved by solving the following:
\begin{align}
    \min_{\mathbf{s}^1:|\mathbf{s}^1| \leq b} \max_i \min_{j\in\mathbf{s}^1\cup\mathbf{s}^0} \Delta(x_i,x_j)    
\end{align}

where $\Delta$ denotes some distance metric, $\mathbf{s}^1$ denotes the chosen AL batch, and $\mathbf{s}^0$ denotes the labeled set. The authors of~\cite{sener2018active} use a robust $k$-center algorithm to solve this objective. The points selected in this fashion are thus diverse and representative of $\mathcal{D_{UL}}$, allowing for future training to produce $f$ that generalizes better on the domain. In our experiments, we choose to represent each $x$ by its penultimate layer embedding $f_{L-1}(x)$ when calculating $\Delta(x_i,x_j)$.\looseness-1

\textbf{FASS}~\cite{pmlr-v37-wei15} applies a two-step process of uncertainty sampling and submodular selection. Specifically, the top $b\times k$ most uncertain $x\in \mathcal{D_{UL}}$ are initially chosen and stored in $\mathcal{U}$, where $k$ is a configurable hyper-parameter. Next, a submodular~\cite{iyer2021submodular} set function $r:2^\mathcal{U}\rightarrow \mathbb{R}$ is instantiated on $\mathcal{U}$. Submodular function maximization subject to cardinality constraints is known to find diverse or representative subsets; thus, \textbf{FASS}~\cite{pmlr-v37-wei15} selects $\mathbf{s}^1\subseteq\mathcal{U}$ such that $|\mathbf{s}^1|=b$ using submodular maximization with cardinality constraints on $r$, thereby capturing both uncertainty (via $\mathcal{U}$) and diversity (via $r$) in $\mathcal{D_{UL}}$. In our experiments, we use \textbf{entropy sampling}~\cite{settles.tr09} for the uncertainty measure and \textbf{facility location}~\cite{iyer2021submodular} for $r$. As before, we use penultimate layer embeddings given by $f_{L-1}(x)$ for each $x\in\mathcal{U}$ when instantiating $r$.\looseness-1

\textbf{\textsc{Badge}}~\cite{ash2020deep}, a very recent approach free of hyper-parameter tuning, first computes the last linear layer gradients of the hypothesized loss on each of the instances in $\mathcal{D_{UL}}$. Specifically, if we denote $\hat{y}(x)=\max_i f_L(x)_i$ and $\mathcal{L}$ as the loss used, then for each $x\in\mathcal{D_{UL}}$, we compute $g(x) = \nabla_{\theta_{L-1}}\mathcal{L}(\hat{y}(x),f_L(x))$. Denoting $\mathcal{G}=\{g(x)|x\in\mathcal{D_{UL}}\}$, \textbf{\textsc{Badge}}~\cite{ash2020deep} then runs the \textsc{k-means++} initialization algorithm~\cite{1283494} on $\mathcal{G}$ to obtain $b$ centers, which are likely to all have large loss gradient magnitudes (which captures uncertainty brought about by large changes to the loss) and diverse loss gradient directions (which captures diversity brought about by diverse parameter updates). The points corresponding to these $b$ centers are then chosen as the AL batch by \textbf{\textsc{Badge}}~\cite{ash2020deep}, which helps future training on $f$ better adapt to the domain.\looseness-1

Lastly, \textbf{\textsc{GLISTER-ACTIVE}}~\cite{killamsetty2020glister} selects points by solving the following bi-level optimization problem:

\begin{align}
    \argmax_{\mathbf{s}^1\subseteq \mathcal{D_{UL}},|\mathbf{s}^1|\leq b} LL(\argmax_\theta LL(\theta,P(\mathbf{s}^1)),\mathcal{V})
\end{align}

where $LL(\theta,S)$ denotes the log-likelihood on the set $S$ using $f$ parameterized by $\theta$, $P(S) = \{(x,\hat{y}(x))|x\in S\}$, and $\mathcal{V}$ denotes a validation set of interest. In our experiments, we set $\mathcal{V}$ as the union of the labeled set and $\mathcal{D_{UL}}$ with hypothesized labels (\emph{e.g.}, the set $P(\mathcal{D_{UL}})$). Here, the inner optimization problem learns model parameters on the pseudo-labeled AL batch, and the outer optimization problem selects a set of unlabeled instances from $\mathcal{D_{UL}}$ that maximizes the log-likelihood with respect to the validation set using the trained model parameters from the inner problem. The authors of~\cite{killamsetty2020glister} use an online meta approximation algorithm to carry out the bi-level optimization. Ultimately, \textbf{\textsc{GLISTER-ACTIVE}}~\cite{killamsetty2020glister} selects $\mathbf{s}^1$ that is representative of the domain (since we use our specific choice of $\mathcal{V}$), which aids in producing $f$ that generalizes better on the domain.\looseness-1 

Since the aim of deep AL methods is to reduce the labeling cost, AL is therefore most useful when considered in large data settings. Techniques such as \textsc{BALD}~\cite{houlsby2011bayesian}, adversarial-based AL techniques~\cite{ducoffe2018adversarial}, and \textsc{VAAL}~\cite{sinha2019variational} are not as scalable; therefore, we do not consider them in most of our experiments in our main sections. For larger AL batch sizes, \textsc{BALD} requires intractably large collections of Monte Carlo samples for its Bayesian estimates, while \textsc{VAAL}~\cite{sinha2019variational} requires maintaining and training three models -- the main supervised model, the VAE model, and the discriminator~\cite{sinha2019variational}. Additionally, other adversarial methods such as those studied in~\cite{ducoffe2018adversarial} require expensive perturbations to perform AL selection. For completeness, we perform a few small-scale experiments to compare all AL baselines (including \textsc{BALD}, \textsc{VAAL}, and other adversarial methods) in \hyperref[sec:a2]{Appendix B}. Specifically, we contrast the running times, test performance, and labeling efficiencies of all these AL approaches.

\section{Experiments}
\begin{figure} 
    \centering
    \includegraphics[width=0.95\linewidth]{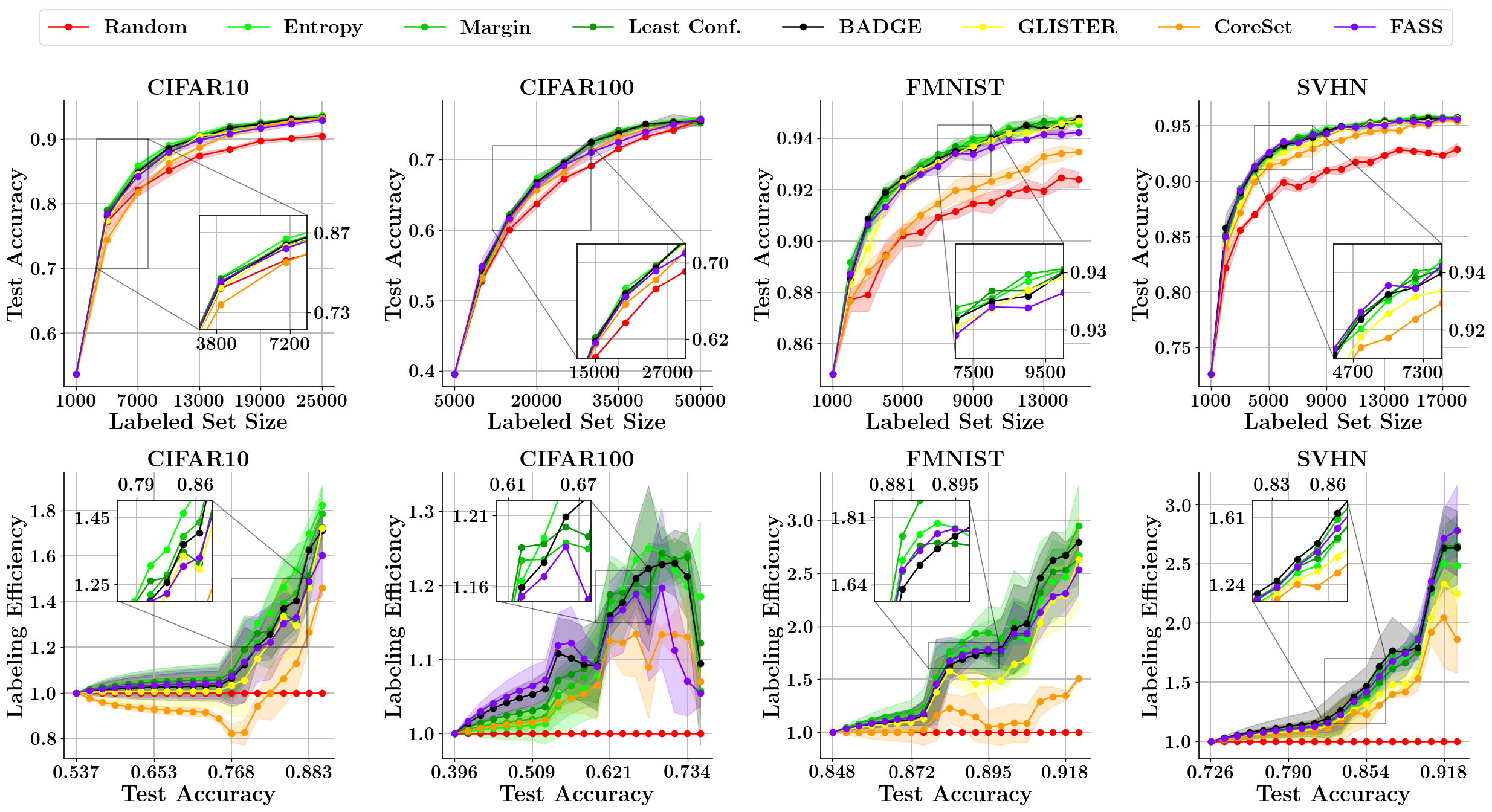}
    \vspace{-2ex}
    \caption{Baseline results on the CIFAR-10, CIFAR-100, Fashion-MNIST, and SVHN. We observe that most AL schemes achieve labeling efficiencies ranging from $1.3\times$ to $3\times$ compared to RS.}
    \label{fig:baseline}
\end{figure}

In this section, we report experiments and results to study the previously listed facets of effective evaluation (\hyperref[sec:facets]{\textbf{Sec. 2}}) using {\color{blue}\href{https://github.com/decile-team/distil}{DISTIL}} as discussed earlier in this paper. Furthermore, we use a common set of experimental parameters unless otherwise specified, which are detailed in \hyperref[sec:a1]{Appendix A}. Our data augmentations consist of random horizontal flips and random croppings ($p=0.5$) of the images padded by 4 pixels on each side, followed by data normalization. These augmentations occur every time a data instance is retrieved during training. To encourage reproduction of our results, we provide simple instructions to execute our experiments and list the compute resources needed to run these experiments in \hyperref[sec:a1]{Appendix A}. We evaluate our algorithms on the following five image classification data sets: CIFAR-10~\cite{Krizhevsky09learningmultiple}, CIFAR-100~\cite{Krizhevsky09learningmultiple}, MNIST~\cite{lecun-98}, Fashion MNIST~\cite{xiao2017fashionmnist}, and SVHN~\cite{svhncite}. For each experiment, we initialize each AL strategy with an identical initial model and initial labeled seed set. Each model is trained until either the training accuracy reaches 99\%, the epoch count reaches 300, or the training accuracy does not improve after 10 epochs. On all data sets except MNIST~\cite{lecun-98}, a ResNet-18~\cite{he2015deep} and a VGG-11~\cite{simonyan2014very} are used. On MNIST~\cite{lecun-98}, we use a model similar to a LeNet~\cite{lecun-98} (see \hyperref[sec:a1]{Appendix A}). Lastly, we run each experiment three times and report the average across all runs and the associated variance. We allude to significance test results here and provide details of the same in  \hyperref[sec:a7]{Appendix G}.

\label{sec:e1}
\textbf{4.1 Baseline Experiments \& Labeling Efficiency.} To have an accurate comparison of AL selection algorithms, we first conducted a baseline experiment across multiple data sets to gauge the effectiveness of each algorithm in a general scenario. As highlighted in \hyperref[sec:al-algos]{\textbf{Sec. 3}}, we restrict our comparison to random sampling, entropy sampling, margin sampling, least confidence sampling, \textsc{Badge}, \textsc{GLISTER-ACTIVE}, \textsc{CoreSet}, and \textsc{FASS}. Other algorithms are compared in \hyperref[sec:a2]{Appendix B}. In Figure~\ref{fig:baseline} (upper plots), we plot the obtained test accuracy as a function of the size of the labeled set. In addition, in the lower plots, the labeling efficiency of each selection algorithm with respect to random sampling is displayed. Specifically, the labeling efficiency for a given test accuracy is computed as the ratio of the minimum number of labeled training instances needed for an AL algorithm to reach that accuracy to the minimum number of labeled training instances needed by random sampling to reach that accuracy. The results in Figure~\ref{fig:baseline} are obtained using a ResNet-18 model on CIFAR-10, CIFAR-100, SVHN, and Fashion-MNIST. The results for VGG-11's performance and the results for the MNIST data set are in \hyperref[sec:a2]{Appendix B}. In each baseline, \emph{we observe that the top-performing strategies exhibit labeling efficiencies that range from $1.3\times$ (CIFAR-100) to $3\times$ (SVHN and FMNIST), indicating that the commonly used AL selection algorithms achieve their intended purpose of reducing labeling costs}. For MNIST (\hyperref[sec:a2]{Appendix B}), we observe a labeling efficiency of $4.5\times$. This addresses \hyperref[sec:p1]{\textbf{P1}} (unified experimental setting) and \hyperref[sec:p2]{\textbf{P2}} (labeling efficiency).


\begin{figure}
    \centering
    \includegraphics[width=0.95\linewidth]{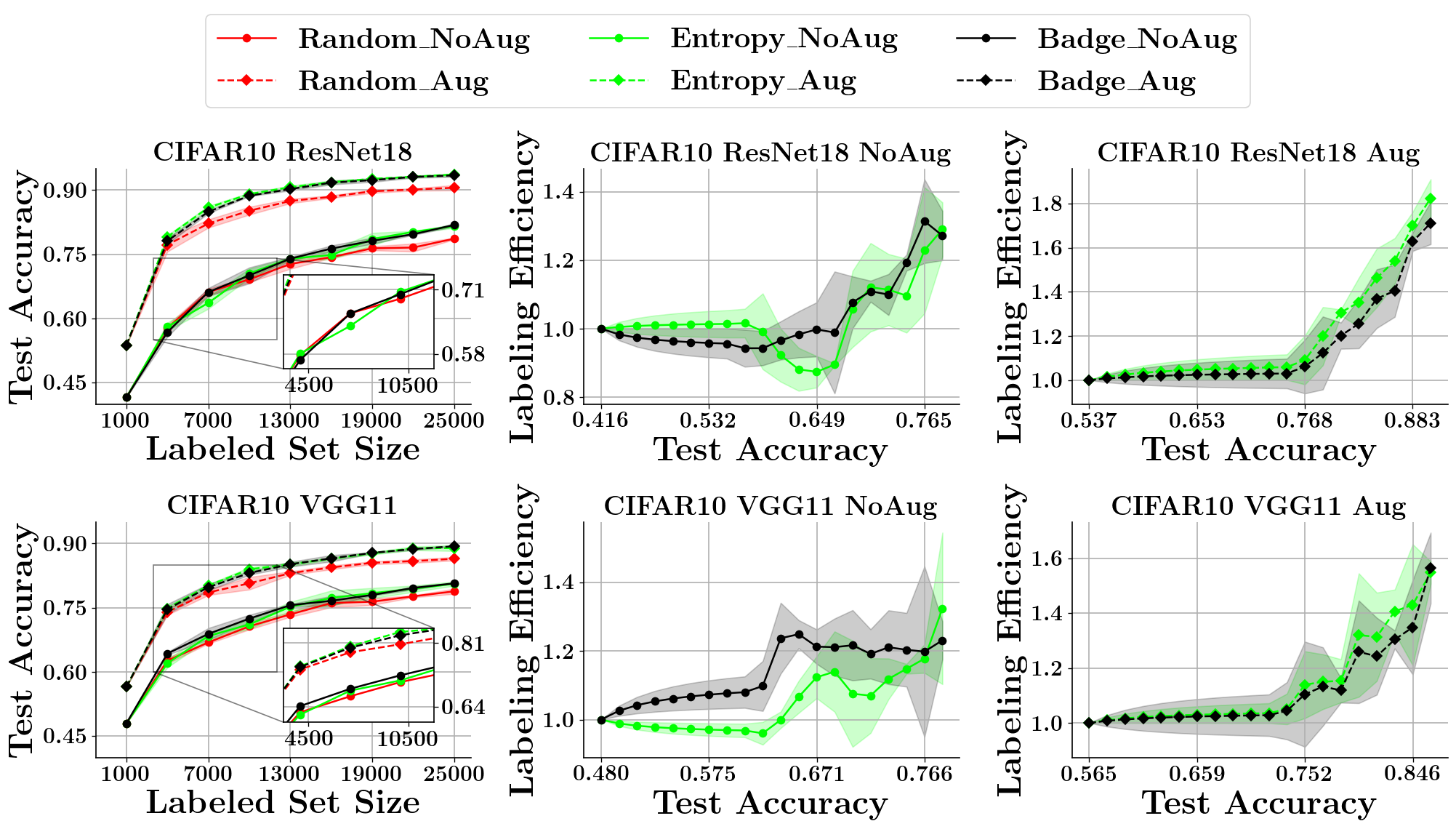}
    \vspace{-2ex}
    \caption{Comparing AL performance of ResNet-18 (top) and VGG-11 (bottom) on CIFAR-10 with and without augmentation. Data augmentation not only increases test accuracy but also improves the labeling efficiencies of AL. Furthermore, \textsc{Badge} outperforms entropy sampling without data augmentation, but \textsc{Badge} loses its advantage over entropy sampling when data augmentation is used.\looseness-1}
    \label{fig:augmentation}
\end{figure}

\label{sec:e2}
\textbf{4.2 Data Augmentation.} To better understand the effect of data augmentation on image data sets in an AL setting, we compare AL on CIFAR-10~\cite{Krizhevsky09learningmultiple} with and without the use of data augmentation. We do this for \textsc{Badge} and entropy sampling, which are the best-performing strategies in Figure~\ref{fig:baseline} and are also representative techniques among uncertainty and diversity-based approaches. The results are shown in Figure~\ref{fig:augmentation}. First, the use of augmentation \emph{significantly and consistently improves the accuracy by almost 10\% when compared to no augmentation}. 
\begin{wrapfigure}{r}{0.5\textwidth}
    \vspace{-2ex}
    \includegraphics[width=0.5\textwidth]{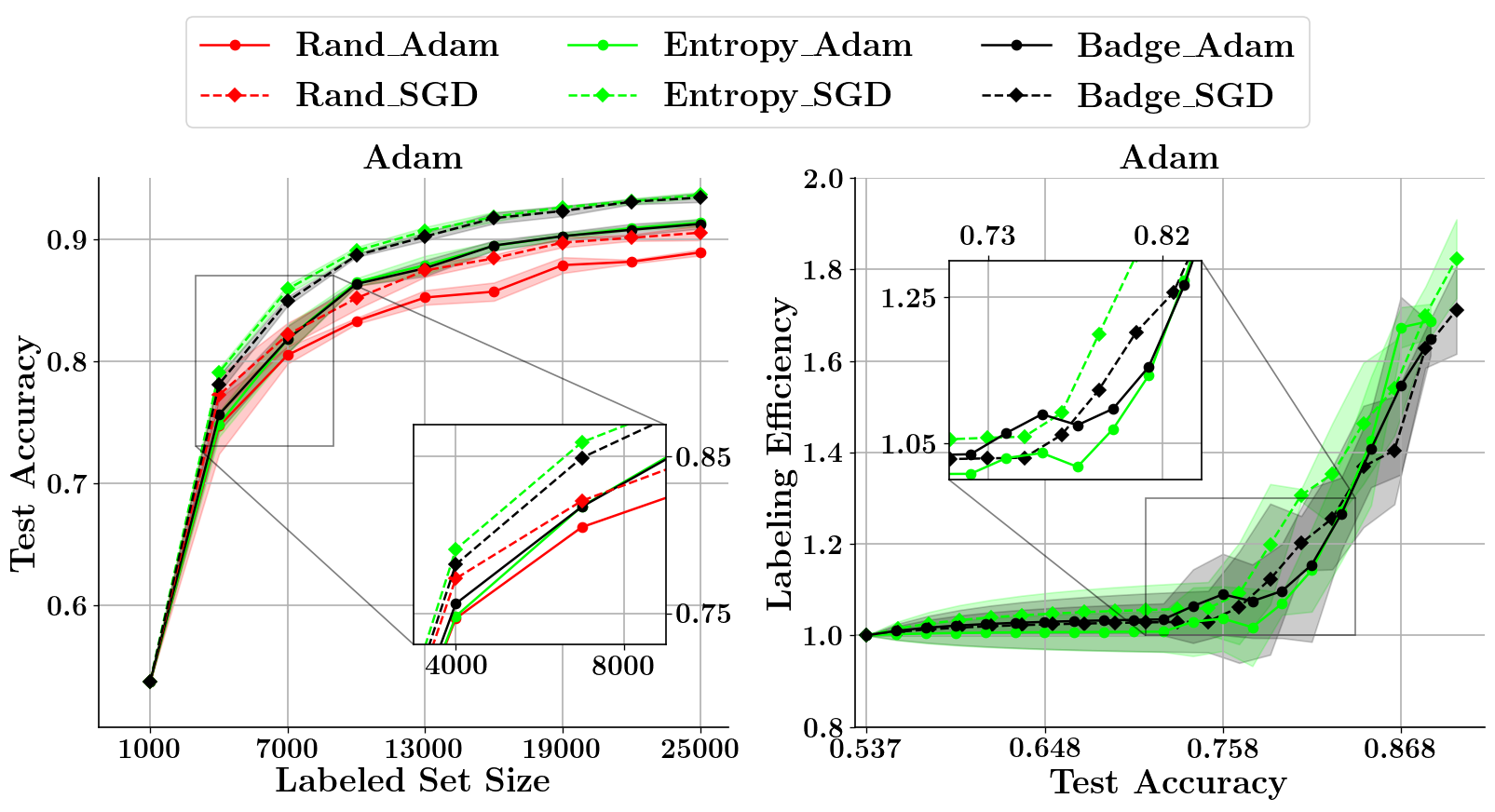}
    \vspace{-3ex}
    \caption{Adam v. SGD. SGD achieves a higher final labeling efficiency and maintains higher generalization performance compared to Adam.}
    \label{fig:adam}
\end{wrapfigure}
Secondly, 
\emph{augmentation also increases the labeling efficiency of AL}. The labeling efficiency improves from $1.4\times$ to $1.8\times$ for ResNet-18 and $1.3\times$ to $1.7\times$ for VGG-11 towards the end. Thirdly, we note that while \textsc{Badge} actually outperforms entropy sampling in the no data augmentation case (which corroborates the results shown in~\cite{ash2020deep}), \emph{there is no significant difference between the two with the addition of data augmentation} (entropy sampling performs better than BADGE in just one of the selection rounds with $\alpha=0.05$). These results can be partly explained by the fact that data augmentation naturally introduces more diversity into the training. This addresses \hyperref[sec:p3]{\textbf{P3}}.\looseness-1

\label{sec:e3}
\textbf{4.3 Optimizer and Other Generalization Techniques.} Another aspect of AL evaluation involves analyzing the effect of the training strategy used. Past work in AL has mostly used adaptive gradient methods for optimization,  such as Adam~\cite{ash2020deep,sinha2019variational} and RMSProp~\cite{sener2018active}. Given our analysis on generalization performance, it is natural to then consider the effect of using SGD over Adam since a number of works have argued that deep neural networks generalize better when trained with SGD as opposed to Adam~\cite{wilson2017marginal,zhou2020theoretically}. We study this in the context of AL and with the use of data augmentation. In particular, we run our experiment on CIFAR-10~\cite{Krizhevsky09learningmultiple} with ResNet-18 and compare Adam against SGD. For SGD, we use a momentum of 0.9, a weight decay of $0.0005$, and a cosine-annealing learning rate ($T_{max} = 300$). For Adam, we use no weight decay and the typical choice of $\beta_1, \beta_2$ as 0.9 and 0.999, respectively. For both, we use a learning rate of 0.01. Based on our results (Figure~\ref{fig:adam}), \emph{we see that Adam results in reduced test accuracy and lower labeling efficiency compared to SGD; however, it converges faster than SGD} (for example, entropy sampling with SGD scores higher than with Adam in all rounds except the initial round with $\alpha=0.05$). Hence, the observations from~\cite{wilson2017marginal,zhou2020theoretically} seem to also apply in the AL setting. Finally, we study the effect of some more advanced generalization approaches, {\em viz.}, stochastic weight averaging (SWA)~\cite{izmailov2019averaging} and shake-shake regularization~\cite{gastaldi2017shakeshake} in \hyperref[sec:a3]{Appendix C}, and we see that the takeaways are consistent: \emph{These approaches all improve the test performance of AL while maintaining high labeling efficiency}. This addresses \hyperref[sec:p4]{\textbf{P4}}.\looseness-1
\begin{figure}
    \centering
    \includegraphics[width=0.95\linewidth]{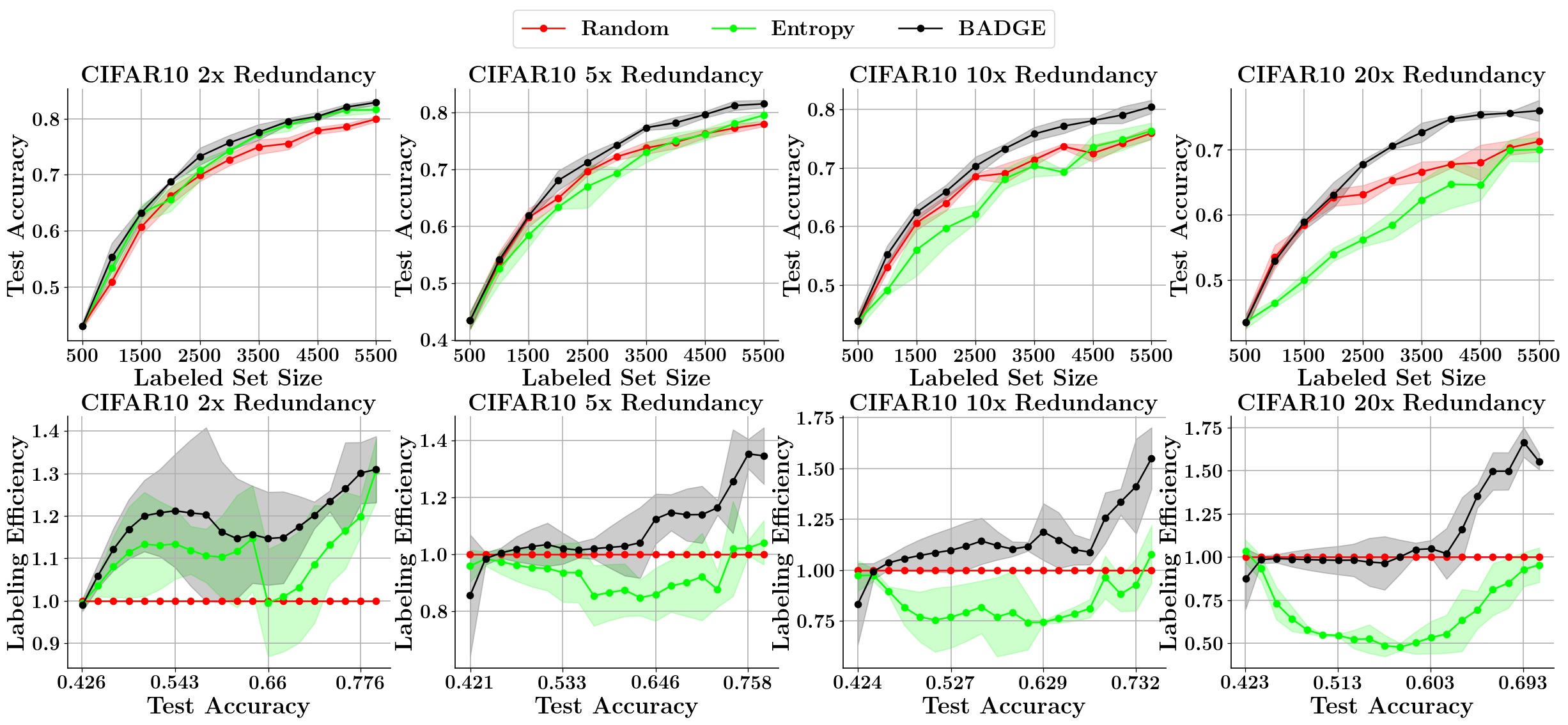}
    \vspace{-2ex}
    \caption{AL results by varying the amount of redundancy in CIFAR-10. Larger amounts of redundancy affect the performance of entropy sampling more than \textsc{Badge}.}
    \label{fig:redundancy}
\end{figure}

\label{sec:e4}

\textbf{4.4 Active Learning with Redundancy.} AL strategies are studied on academic data sets that are often not representative of large-scale data sets in real-world applications. Such data sets often contain many redundant data instances. To effectively evaluate these AL strategies, we rerun our baseline setting on a variant of CIFAR-10~\cite{Krizhevsky09learningmultiple} where a set of unique data instances is kept fixed while another disjoint set of data instances is duplicated until the original size of 50k training instances is reached. We construct this data set such that the original class ratios are not altered. Consequently, this allows us to evaluate the effect of redundancy under the scope of \hyperref[sec:p5]{\textbf{P5}}. In all our experiments, we keep 5k unique data instances fixed. Figure \ref{fig:redundancy} shows the results for this setting when the duplicated set is repeated $2\times, 5\times, 10\times,$ and $20\times$. \emph{We see that entropy sampling~\cite{settles.tr09} progressively does worse when more redundancy is introduced whereas \textsc{Badge}~\cite{ash2020deep} exhibits robustness to redundancy}. Furthermore, the more redundancy there is, the more the labeling efficiency of \textsc{Badge} increases towards the end. We summarize by concluding our observation in \hyperref[sec:p5]{\textbf{P5}}: \emph{Diversity-based methods are robust against redundant data while uncertainty-based methods poorly handle redundant data}. In \hyperref[sec:a4]{Appendix D}, we repeat this experiment by generating redundancies using augmentations on existing instances (random cropping and translation) as opposed to simple duplication -- we observe the same takeaways as above.\looseness-1

\begin{figure}
    \centering
    \includegraphics[width=0.95\linewidth]{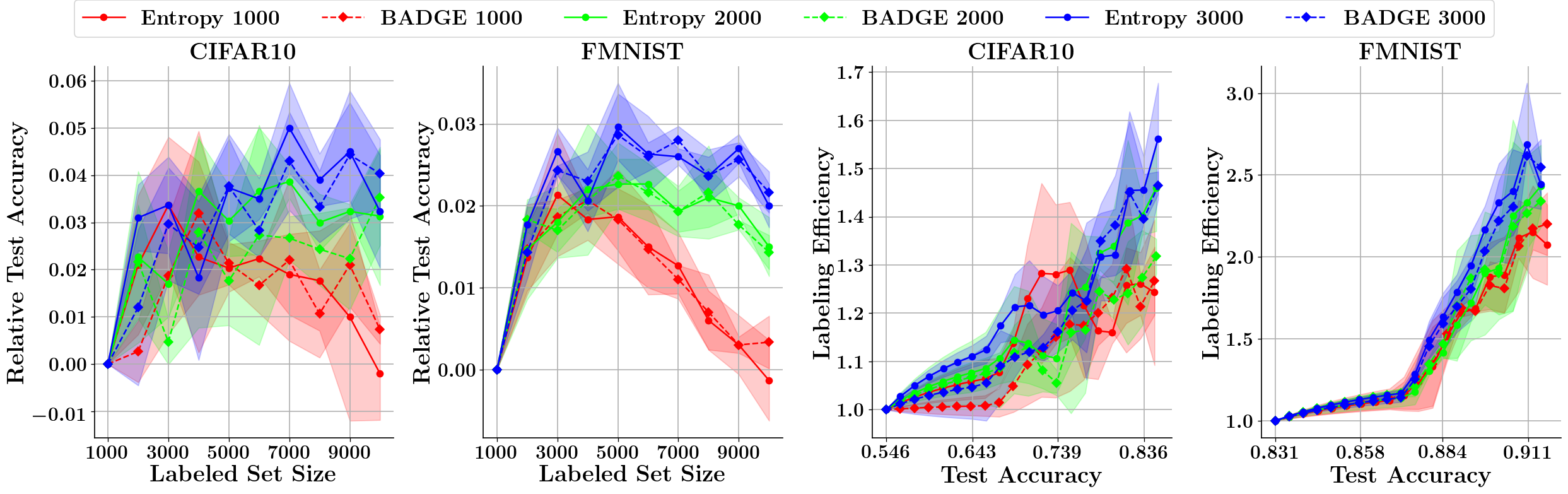}
    \vspace{-2ex}
    \caption{AL results by varying the number of available instances per class in the unlabeled portion of CIFAR-10. The left two plots show relative test accuracies (compared to RS) while the right two show labeling efficiencies. 1000, 2000, and 3000 in the legend correspond to the number of examples per class in the unlabeled set. The more the number of examples there are per class, the better the relative test accuracy \& labeling efficiency of AL is compared to RS.}
    \label{fig:alvrs}
\end{figure}

\label{sec:e5}
\textbf{4.5 Effect of Examples Per Class.} Since the labeling efficiencies achieved in CIFAR-100 ($1.3\times$) are much lower than those achieved in CIFAR-10 ($1.8\times$), we hypothesize that the number of examples per class in the unlabeled set plays an important role in how well AL works compared to RS. We suspect that having more examples per class in the unlabeled set allows for AL to make better choices in refining class performance. To verify this hypothesis, we conduct an experiment on CIFAR-10 and Fashion-MNIST wherein we vary the number of unlabeled data instances per class from 1k up to 3k. Results for this experiment are presented in Figure~\ref{fig:alvrs}. We see that \emph{having a higher number of data instances per class results in higher relative test accuracies and labeling efficiencies with respect to random sampling}. Hence, scenarios in which the unlabeled set is suspected to have few instances per class may be susceptible to reduced AL benefit as stated in \hyperref[sec:p6]{\textbf{P6}} and as illustrated in this experiment.

\begin{table}
\centering
\small{
\begin{tabulary}{\textwidth}{|l*{5}{|c}|}
\hline \textbf{Alg\_Budget} & \textbf{1000 points} & \textbf{7000 points} & \textbf{13000 points} & \textbf{19000 points} & \textbf{25000 points} \\
\hline \textsc{Rand}\_1000 & $55.6 \pm 0.0$ & $82.5 \pm 0.5$ & $87.8 \pm 0.6$ & $89.1 \pm 0.8$ & $91.0 \pm 0.2$ \\
\textsc{Rand}\_3000 & $55.6 \pm 0.0$ & $82.2 \pm 1.2$ & $87.0 \pm 0.2$ & $89.9 \pm 0.5$ & $90.5 \pm 0.5$ \\
\textsc{Rand}\_6000 & $55.6 \pm 0.0$ & $82.0 \pm 0.8$ & $87.7 \pm 0.7$ & $89.4 \pm 0.7$ & $90.7 \pm 0.6$ \\ 
\hline \textsc{Ent}\_1000 & $55.6 \pm 0.0$ & $84.3 \pm 0.9$ & $88.6 \pm 1.2$ & $91.9 \pm 0.9$ & $92.8 \pm 0.8$ \\
\textsc{Ent}\_3000 & $55.6 \pm 0.0$ & $83.3 \pm 1.4$ & $89.4 \pm 1.2$ & $91.7 \pm 0.8$ & $92.5 \pm 0.8$ \\
\textsc{Ent}\_6000 & $55.6 \pm 0.0$ & $82.7 \pm 1.0$ & $88.5 \pm 0.1$ & $91.0 \pm 1.1$ & $92.4 \pm 0.8$ \\
\hline \textsc{Badge}\_1000 & $55.6 \pm 0.0$ & $83.5 \pm 1.5$ & $89.0 \pm 0.6$ & $90.8 \pm 1.2$ & $92.4 \pm 0.6$ \\
\textsc{Badge}\_3000 & $55.6 \pm 0.0$ & $83.4 \pm 1.3$ & $88.4 \pm 0.9$ & $91.5 \pm 0.7$ & $92.6 \pm 0.5$ \\
\textsc{Badge}\_6000 & $55.6 \pm 0.0$ & $82.13 \pm 1.4$ & $88.1 \pm 0.6$ & $91.1 \pm 0.9$ & $92.3 \pm 1.1$ \\ \hline
\end{tabulary}}
\vspace{1ex}
\caption{Test accuracy and standard deviation on CIFAR-10 across varying AL batch sizes. We find that reasonable choices of the batch size make little difference in the achieved test accuracies.}
\label{tab:budget}
\vspace{-4ex}
\end{table}

\begin{table}
\vspace{-1ex}
\centering
\small{
\begin{tabulary}{\linewidth}{|l*{5}{|c}|}
\hline \textbf{Alg\_Init} & \textbf{1000 points} & \textbf{4000 points} & \textbf{7000 points} & \textbf{10000 points} & \textbf{25000 points} \\ \hline
\textsc{Ent}\_FL & $57.8 \pm 0.0$ & $78.6 \pm 0.7$ & $85.8 \pm 0.8$ & $88.2 \pm 0.8$ & $93.5 \pm 0.2$ \\
\textsc{Ent}\_Rand & $53.7 \pm 0.0$ & $79.1 \pm 0.3$ & $85.9 \pm 0.4$ & $89.0 \pm 0.3$ & $93.6 \pm 0.2$ \\ \hline
\textsc{Badge}\_FL & $57.8 \pm 0.0$ & $77.9 \pm 0.3$ & $85.0 \pm 0.7$ & $88.5 \pm 0.5$ & $93.2 \pm 0.4$ \\
\textsc{Badge}\_Rand & $53.8 \pm 0.0$ & $78.1 \pm 0.7$ & $84.9 \pm 0.4$ & $88.6 \pm 0.1$ & $93.4 \pm 0.4$ \\ \hline 
\end{tabulary}}
\vspace{1ex}
\caption{Test accuracy and standard deviation on CIFAR-10 across different seed set initializations. We find that the initial benefit of the carefully constructed seed sets vanishes after only a few rounds.}
\label{tab:seedset}
\vspace{-4ex}
\end{table}

\label{sec:e6}
\textbf{4.6 Effect of Varying AL Batch Sizes and Initializations.} To better understand how the choice of the AL batch size affects the evaluation of AL algorithms, we run an experiment on CIFAR-10 where the AL batch size is varied between 1000, 3000, and 6000 queried instances. The results are presented in Table \ref{tab:budget}. We see that \emph{the AL batch size has little effect on each selection algorithm's obtained test accuracies and labeling efficiencies} (no two configurations of the AL batch size had accuracies significantly different with $\alpha=0.05$ for more than 20\% of the selection rounds). These results show that the choice of AL batch size is not heavily linked with an algorithm's labeling efficiency. Thus, the choice of the AL batch size can be made to optimize the labeling-training pipeline in real-world settings.\looseness-1 

Almost all AL strategies assume a warm start; {\em i.e.}, a small labeled seed set and a model trained on this set are already given. We next study the effect of choosing a sophisticated seed set on the generalization performance and labeling efficiency of AL. In this experiment, we compare a seed set formed by a facility location selection (to select a representative subset)~\cite{iyer2021submodular,kaushal2019learning} against a randomly chosen seed set by repeating the baseline setting with these seed sets. The results are shown in Table \ref{tab:seedset}. We observe that the carefully constructed seed sets offer a better initial accuracy; however, this benefit is quickly lost after only a few rounds of AL (entropy sampling with a random initialization versus entropy sampling with a facility location initialization is the only comparison to show 
a significant difference in test accuracy in a selection round with $\alpha=0.05$). \emph{We conclude that the seed set initialization does play a role very early on, but its influence diminishes after a few rounds of AL} as stated in \hyperref[sec:p7]{\textbf{P7}}.\looseness-1


\label{sec:e7}
\textbf{4.7 Model Reset {\em vs.} Update.} We now analyze a strategy for reducing the training time wherein we update (fine-tune) the model from the previous rounds of AL. We perform this experiment on CIFAR-10, and we compare the update strategy with the resetting strategy. We see that the initial test accuracy is worse for every AL strategy when the model is updated in every round as opposed to when it is reset at every round. This is also corroborated by~\cite{ash2020warmstarting}, wherein warm-starting deep models has the propensity to hurt generalization performance. \emph{However, we interestingly observe that the resetting strategy as well as the updating strategy converge to similar test accuracies at later rounds, implying that it does not matter as much whether we reset or update after a few rounds} (there is not a statistically significant difference in test accuracy with $\alpha=0.05$ between resetting or updating for any given strategy in the later rounds). 
\begin{wrapfigure}{r}{0.5\textwidth}
    \vspace{-2ex}
    \includegraphics[width=0.5\textwidth]{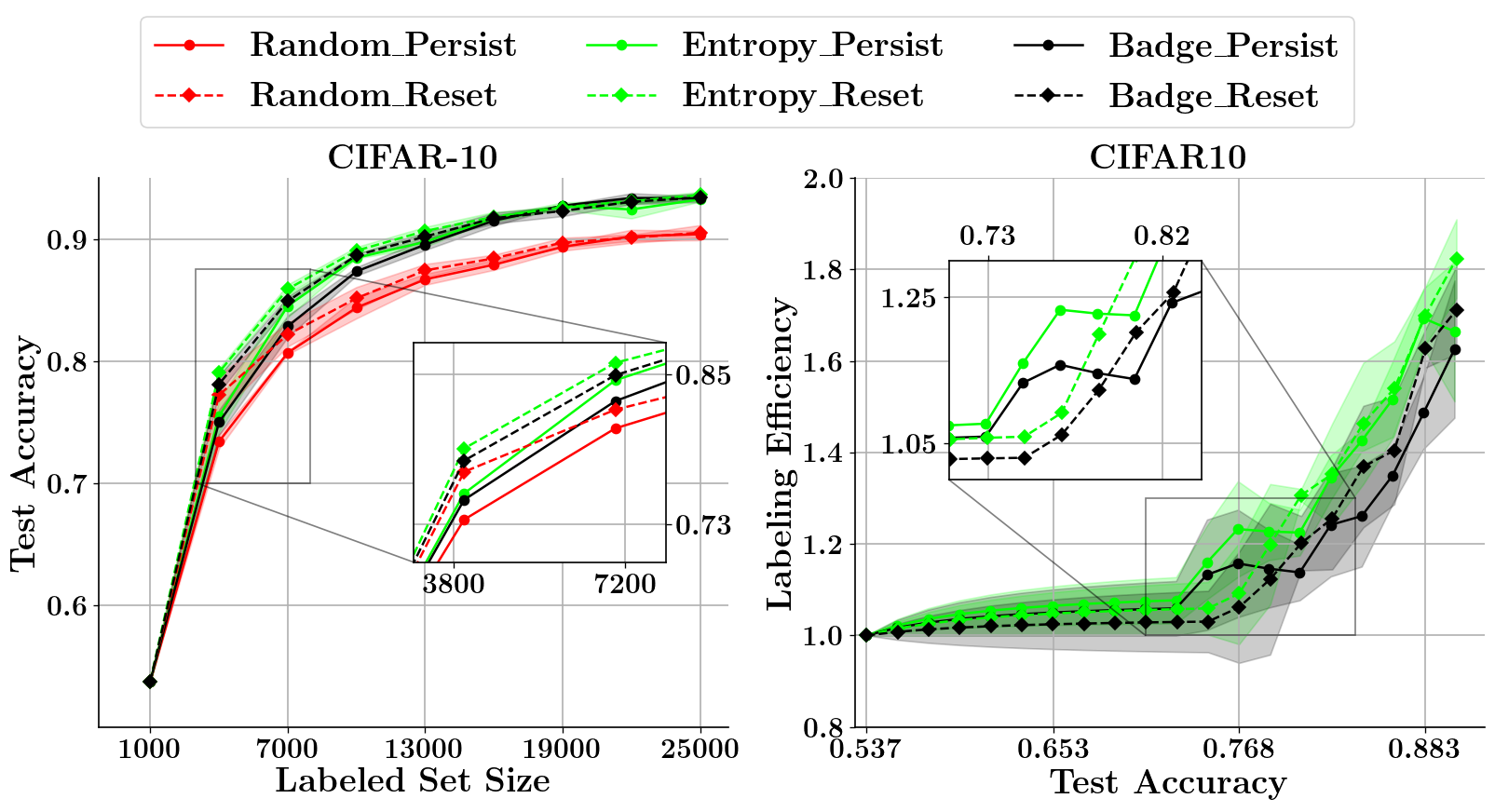}
    \vspace{-3ex}
    \caption{Reset v. Update. Resetting the model after each AL round achieves better initial performance, but both strategies converge later on.}
    \label{fig:persist}
\end{wrapfigure}
A critical difference between our setup and that of~\cite{ash2020warmstarting} is that the latter does not use data augmentation. Furthermore, the latter employs Adam instead of SGD. In \hyperref[sec:a5]{Appendix E}, we report an experiment identical to the one conducted in this section that employs Adam but without data augmentation. In this setting, we find that updating the model is consistently worse than resetting the model in every round. This illustrates another interesting benefit of data augmentation: we do not significantly lose generalization performance if we choose to update the model, which can save compute resources for expensive training tasks. In \hyperref[sec:a5]{Appendix E}, we illustrate this point on other data sets.\looseness-1   

\label{sec:e8}
\textbf{4.8 Scalability, Running Times, and Compute.} Next, we discuss the running times of AL. Repeating the baseline, we run AL with an AL batch size of 3k instances for 8 rounds on CIFAR-10 up to a total of 25k instances (half of CIFAR-10). The total time taken for the intermediate training (across rounds) is 8 hours; in contrast, the total time taken by \textsc{Badge} selection (across rounds) is 1 hour whereas entropy sampling takes 3 minutes. The training time almost triples when we run for the entire data set. Specifically, the increase in training time over previous work can largely be attributed to data augmentation and SGD, which causes slower model convergence. This further implies increased energy consumption and CO2 emissions. To help alleviate and reduce the training times and compute, we suggest a few approaches. First, we can increase the AL batch size, thereby reducing the number of training rounds (which we have shown to be a significant computational bottleneck). Secondly, we can update the model from the previous rounds. This has negligible impact on performance in later rounds (as shown in \hyperref[sec:e7]{Sec. 4.7}) but only speeds up the running time by ~$1.07\times$. As a faster training strategy, we also examine the use of data subset selection via the recently proposed \textsc{Grad-Match}~\cite{killamsetty2021gradmatch} algorithm. Using only 30\% of the labeled data for training, we see that there is negligible loss in accuracy while achieving a further $3\times$ speedup in training. Finally, since entropy sampling is $20\times$ faster than \textsc{Badge}, we propose the use of entropy sampling unless the data set has redundancy and repetitions. More details on the timings are in \hyperref[sec:a6]{Appendix F}.

\section{Conclusion} 
In this work, we study AL in the context of  recent deep learning advances like data augmentation and so forth. We  provide several insights for AL practitioners. Summarily: \textbf{1)} It is beneficial to use a training loop which applies data augmentation, SGD, and other AL generalization approaches to improve test accuracy and labeling efficiency. \textbf{2)} We study the labeling efficiency of AL across multiple image classification data sets. \textbf{3)} Sophisticated diversity-based approaches like \textsc{Badge}  make sense only if the data has a lot of redundancy; uncertainty sampling is otherwise good enough and more compute-efficient. \textbf{4)} Having more examples per class in the unlabeled set leads to higher labeling efficiencies from AL. \textbf{5)} Reasonable choices of the AL batch size and seed set do not affect AL performance. \textbf{6)} Updating the model from the previous round is comparable to retraining from scratch in the later AL rounds. \textbf{7)} Approaches like model updating and data subset selection can significantly reduce the turnaround time and carbon emissions from AL. The last point is critical because AL generalization using data augmentation and SGD can significantly decelerate training convergence 
to get desirable accuracy gains. 
We believe that  model updating and data subset selection can significantly reduce these compute costs 
while maintaining comparable accuracy. As future work, we plan to expand our study of AL evaluation beyond image classification tasks via our {\color{blue}\href{https://github.com/decile-team/distil}{DISTIL}} toolkit, analyzing the same facets in the context of object detection, NLP, and so forth.

\newpage
\bibliographystyle{plain}
\bibliography{main}

\newpage
\appendix

\begin{center}
\part{Supplementary Material for Effective Evaluation of Deep Active Learning on Image Classification Tasks} 
\end{center}
\parttoc 

\label{sec:a1}
\section{Reproducibility and Experimental Details}

\subsection{Provided Experiments}

In {\color{blue}\href{https://github.com/decile-team/distil}{DISTIL}}, we provide our experiments as Jupyter notebooks that are intended to be executed on Google Colab GPU VMs. In our executions, we utilized Colab Pro benefits in procuring these VMs. These VMs typically feature one 16 GB VRAM variant of the Nvidia P100 or V100 GPU, 12 GB to 25 GB of RAM, and two Intel(R) Xeon(R) CPUs. By using Google Colab VMs, we provide a widely accessible means of easily reproducing our results. Indeed, running each experiment amounts to running each cell within the Google Colab VM.

In our experiments, we also strive to use the same initial model whenever possible. In most of the provided Jupyter notebooks, we provide a configurable option to train a new initial model or to download and use a model from a Google Drive link. By default, we configure each experiment to train a new initial model. To then use that initial model in future runs, we recommend storing the model in a different location in Google Drive and downloading it from a shareable link as previously described.

Lastly, our notebooks provide a means of checkpointing since Colab VMs are not indefinitely kept alive. If an experiment is interrupted in this fashion, then simply running the preparation cells followed by the desired experiment cell should resume the experiment from the last checkpoint.

\subsection{Code Repositories and Licenses}

Our experiments utilize our previously mentioned unified framework that draws upon contributions from existing sources and data sets. The following repositories and data sets are used, and their licenses are also provided:

\begin{itemize} 
    \item \href{https://pytorch.org/}{{\color{blue}PyTorch}}~\cite{NEURIPS2019_9015}: Modified BSD
    \item \href{https://github.com/ej0cl6/deep-active-learning}{{\color{blue}Deep Active Learning}}~\cite{DeepALRepo}: None Listed
    \item \href{https://github.com/JordanAsh/badge}{{\color{blue}BADGE}}~\cite{badgerepo}: None Listed
    \item \href{https://github.com/sinhasam/vaal}{{\color{blue}VAAL}}~\cite{sinha2019variational}: BSD 2-Clause License
    \item \href{https://github.com/LTS4/DeepFool}{{\color{blue}DeepFool}}~\cite{Moosavi-Dezfooli_2016_CVPR}: None Listed
    \item \href{https://github.com/jmschrei/apricot}{{\color{blue}Apricot}}~\cite{schreiber2020apricot}: MIT License
    \item \href{https://www.cs.toronto.edu/~kriz/cifar.html}{{\color{blue}CIFAR-10}}~\cite{Krizhevsky09learningmultiple}: MIT License
    \item \href{https://www.cs.toronto.edu/~kriz/cifar.html}{{\color{blue}CIFAR-100}}~\cite{Krizhevsky09learningmultiple}: MIT License
    \item \href{http://yann.lecun.com/exdb/mnist/}{{\color{blue}MNIST}}~\cite{lecun-98}: Creative Commons Attribution-Share Alike 3.0
    \item \href{https://github.com/zalandoresearch/fashion-mnist}{{\color{blue}FMNIST}}~\cite{xiao2017fashionmnist}: MIT License
    \item \href{http://ufldl.stanford.edu/housenumbers/}{{\color{blue}SVHN}}~\cite{svhncite}: CC0 1.0 Public Domain
\end{itemize}

\subsection{Experimental Parameters}

Here, we list in detail the parameters used in each experiment. In our tables, \textbf{Max Epoch} refers to the maximum epoch allowed during training, \textbf{LR} refers to the learning rate used in training, \textbf{Opt. B. Size} refers to the batch size used by the optimizer, \textbf{Train Cut} refers to the training accuracy cutoff, \textbf{Opt.} refers to the optimizer, \textbf{AL B. Size} refers to the AL batch size, and \textbf{Seed Size} refers to the labeled seed set size.

\textbf{Baselines.} In Table \ref{tab:baseparam}, we provide the parameters for our baseline settings. Each row represents a data set and model architecture pair. For MNIST, our model architecture described in the next section is used. Note that the low-data MNIST setting describes the parameters for all strategies except for VAAL~\cite{sinha2019variational}. The training loop for VAAL is necessarily different as three models are trained; hence, we use the training loop provided in the VAAL repository. Also, we opt to use the default parameters specified in the VAAL repository mentioned previously. Additionally, we add functionality to support MNIST to the VAAL training loop. Lastly, CIFAR-100 uses a partitioned version of \textsc{Badge} wherein the unlabeled set is divided into chunks. Each chunk is then handled by \textsc{Badge} individually, and the chosen instances for each chunk are then merged into the final queried set of instances.

\begin{table}[H]
    \centering
    \scriptsize{
    \begin{tabular}{|l*{7}{|c}|}
        \hline
        \textbf{Baselines} & \textbf{Max Epoch} & \textbf{LR} & \textbf{Opt. B. Size} & \textbf{Train Cut} & \textbf{Opt.} & \textbf{AL B. Size} & \textbf{Seed Size} \\
        \hline
        CIFAR-10 ResNet-18 & 300 & 0.01 & 20 & 0.99 & SGD & 3000 & 1000 \\
        \hline
        CIFAR-10 VGG-19 & 300 & 0.01 & 128 & 0.99 & SGD & 3000 & 1000 \\
        \hline
        CIFAR-10 DLA & 300 & 0.01 & 20 & 0.99 & SGD & 3000 & 1000 \\
        \hline
        CIFAR-100 ResNet-18 & 300 & 0.01 & 20 & 0.99 & SGD & 5000 & 5000 \\
        \hline
        MNIST & 300 & 0.01 & 20 & 0.99 & SGD & 1000 & 300 \\
        \hline
        MNIST Low & 3000 & 0.01 & 20 & 0.99 & SGD & 10 & 50 \\
        \hline
        FMNIST ResNet-18 & 150 & 0.01 & 64 & 0.99 & SGD & 1000 & 1000 \\
        \hline
        SVHN ResNet-18 & 150 & 0.01 & 64 & 0.99 & SGD & 1000 & 1000 \\ 
        \hline
        SVHN VGG-19 & 300 & 0.001 & 32 & 0.99 & SGD & 1000 & 1000 \\
        \hline
    \end{tabular}}
    \caption{Experimental parameters used in baseline comparisons. Note: Our baselines use VGG-19, not VGG-11. This is a typo in the main sections of the paper.}
    \label{tab:baseparam}
\end{table}

\textbf{Data Augmentation.} The experiments on data augmentation in \hyperref[sec:e2]{\textbf{Sec. 4.2}} use the following parameters in Table \ref{tab:augparam}. The VGG-11 runs specifically use Adam to match the setting in~\cite{ash2020deep}. Note that an experiment with augmentation and an experiment without augmentation are run for each parameter configuration given in Table \ref{tab:augparam}.

\begin{table}[H]
    \centering
    \scriptsize{
    \begin{tabular}{|l*{7}{|c}|}
        \hline
        \textbf{Exp.} & \textbf{Max Epoch} & \textbf{LR} & \textbf{Opt. B. Size} & \textbf{Train Cut} & \textbf{Opt.} & \textbf{AL B. Size} & \textbf{Seed Size} \\
        \hline
        CIFAR-10 ResNet-18 & 300 & 0.01 & 20 & 0.99 & SGD & 3000 & 1000 \\
        \hline
        CIFAR-10 VGG-11 & 300 & 0.001 & 20 & 0.99 & Adam & 3000 & 1000 \\
        \hline
    \end{tabular}}
    \caption{Experimental parameters used in our data augmentation setting.}
    \label{tab:augparam}
\end{table}

\textbf{Optimizer and Other Generalization Techniques.} The experiments on the choice of optimizer and the effect of other generalization techniques use the following parameters in Table \ref{tab:augopt}. Each SS configuration uses a shake-shake-regularized ResNet-18. Each SWA configuration uses stochastic weight averaging. All other configurations use ResNet-18. All experiments are conducted on CIFAR-10.

\begin{table}[H]
    \centering
    \small{
    \begin{tabular}{|l*{7}{|c}|}
        \hline
        \textbf{Exp.} & \textbf{Max Epoch} & \textbf{LR} & \textbf{Opt. B. Size} & \textbf{Train Cut} & \textbf{Opt.} & \textbf{AL B. Size} & \textbf{Seed Size} \\
        \hline
        SGD & 300 & 0.01 & 20 & 0.99 & SGD & 3000 & 1000 \\
        \hline
        Adam & 300 & 0.01 & 20 & 0.99 & Adam & 3000 & 1000 \\
        \hline
        SWA & 300 & 0.01 & 20 & 0.99 & SGD/SWA & 3000 & 1000 \\
        \hline
        SS & 300 & 0.01 & 20 & 0.99 & SGD & 3000 & 1000 \\
        \hline
        SWA\_SS & 300 & 0.01 & 20 & 0.99 & SGD/SWA & 3000 & 1000 \\
        \hline
    \end{tabular}}
    \caption{Experimental parameters used in our optimizer and other generalization technique experiments.}
    \label{tab:augopt}
\end{table}

\textbf{Redundancy.} The experiments on the effect of redundancy use the following parameters in Table \ref{tab:redparam}. Note that both experiments (duplicating and augmenting) follow the strategy in \hyperref[sec:e4]{\textbf{Sec. 4.4}}, and the following parameter configurations are applicable for each amount of redundancy.

\begin{table}[H]
    \centering
    \small{
    \begin{tabular}{|l*{7}{|c}|}
        \hline
        \textbf{Exp.} & \textbf{Max Epoch} & \textbf{LR} & \textbf{Opt. B. Size} & \textbf{Train Cut} & \textbf{Opt.} & \textbf{AL B. Size} & \textbf{Seed Size} \\
        \hline
        Duplicate & 300 & 0.01 & 20 & 0.99 & SGD & 500 & 500 \\
        \hline
        Augment & 300 & 0.01 & 20 & 0.99 & SGD & 500 & 500 \\
        \hline
    \end{tabular}}
    \caption{Experimental parameters used in our redundancy setting.}
    \label{tab:redparam}
\end{table}

\textbf{Effect of Examples Per Class.} The experiments on the effect of examples per class in the unlabeled set use the following parameters in Table \ref{tab:alvrsparam}. As before, each experiment follows the strategy in \hyperref[sec:e5]{\textbf{Sec. 4.5}}, and the following parameter configurations are applicable for each number of examples per class in the unlabeled set.

\begin{table}[H]
    \centering
    \small{
    \begin{tabular}{|l*{7}{|c}|}
        \hline
        \textbf{Exp.} & \textbf{Max Epoch} & \textbf{LR} & \textbf{Opt. B. Size} & \textbf{Train Cut} & \textbf{Opt.} & \textbf{AL B. Size} & \textbf{Seed Size} \\
        \hline
        CIFAR-10 & 300 & 0.01 & 20 & 0.99 & SGD & 1000 & 1000 \\
        \hline
        FMNIST & 150 & 0.01 & 64 & 0.99 & SGD & 1000 & 1000 \\
        \hline
    \end{tabular}}
    \caption{Experimental parameters used in our examples per class experiment.}
    \label{tab:alvrsparam}
\end{table}

\textbf{AL Batch Sizes and Initializations.} The experiments on the effect of varying AL batch sizes and seed set initializations use the following parameters in Table \ref{tab:budseedparam}. Each experiment uses CIFAR-10 and a ResNet-18 model.

\begin{table}[H]
    \centering
    \small{
    \begin{tabular}{|l*{7}{|c}|}
        \hline
        \textbf{Exp.} & \textbf{Max Epoch} & \textbf{LR} & \textbf{Opt. B. Size} & \textbf{Train Cut} & \textbf{Opt.} & \textbf{AL B. Size} & \textbf{Seed Size} \\
        \hline
        Bud\_1000 & 300 & 0.01 & 20 & 0.99 & SGD & 1000 & 1000 \\
        \hline
        Bud\_3000 & 300 & 0.01 & 20 & 0.99 & SGD & 3000 & 1000 \\
        \hline
        Bud\_6000 & 300 & 0.01 & 20 & 0.99 & SGD & 6000 & 1000 \\
        \hline
        Rand\_Seed & 300 & 0.01 & 20 & 0.99 & SGD & 3000 & 1000 \\
        \hline
        FL\_Seed & 300 & 0.01 & 20 & 0.99 & SGD & 3000 & 1000 \\
        \hline
    \end{tabular}}
    \caption{Experimental parameters used in our AL batch size and seed set initialization settings. Note: FL refers to facility location.}
    \label{tab:budseedparam}
\end{table}

\textbf{Model Reset \textit{vs.} Update.} The experiments on the effect of model resetting versus model updating use the following parameters in Table \ref{tab:resetparam}.

\begin{table}[H]
    \centering
    \scriptsize{
    \begin{tabular}{|l*{7}{|c}|}
        \hline
        \textbf{Exp.} & \textbf{Max Epoch} & \textbf{LR} & \textbf{Opt. B. Size} & \textbf{Train Cut} & \textbf{Opt.} & \textbf{AL B. Size} & \textbf{Seed Size} \\
        \hline
        CIFAR-10 Reset & 300 & 0.01 & 20 & 0.99 & SGD & 3000 & 1000 \\
        \hline
        CIFAR-10 Update & 300 & 0.01 & 20 & 0.99 & SGD & 3000 & 1000 \\
        \hline
        CIFAR-10 N.A. Reset & 300 & 0.01 & 20 & 0.99 & Adam & 3000 & 1000 \\
        \hline
        CIFAR-10 N.A. Update & 300 & 0.01 & 20 & 0.99 & Adam & 3000 & 1000 \\
        \hline
        SVHN Reset & 150 & 0.01 & 64 & 0.99 & SGD & 1000 & 1000 \\
        \hline
        SVHN Update & 150 & 0.01 & 64 & 0.99 & SGD & 1000 & 1000 \\
        \hline
        SVHN N.A. Reset & 150 & 0.01 & 64 & 0.99 & Adam & 1000 & 1000 \\
        \hline
        SVHN N.A. Update & 150 & 0.01 & 64 & 0.99 & Adam & 1000 & 1000 \\
        \hline
    \end{tabular}}
    \caption{Experimental parameters used in our reset versus update settings. N.A. represents an experiment where no augmentation is used.}
    \label{tab:resetparam}
\end{table}

\textbf{DSS.} The experiment using \textsc{GradMatch}~\cite{killamsetty2021gradmatch} uses the following parameters in Table \ref{tab:dssparam}. Again, only 30\% of the data is selected by \textsc{GradMatch}~\cite{killamsetty2021gradmatch} every 35 epochs.

\begin{table}[H]
    \centering
    \small{
    \begin{tabular}{|l*{7}{|c}|}
        \hline
        \textbf{Exp.} & \textbf{Max Epoch} & \textbf{LR} & \textbf{Opt. B. Size} & \textbf{Train Cut} & \textbf{Opt.} & \textbf{AL B. Size} & \textbf{Seed Size} \\
        \hline
        DSS CIFAR-10 & 300 & 0.01 & 20 & 0.99 & SGD & 3000 & 1000 \\
        \hline
    \end{tabular}}
    \caption{Experimental parameters used in our \textsc{GradMatch}~\cite{killamsetty2021gradmatch} experiment.}
    \label{tab:dssparam}
\end{table}

\subsection{MNIST Model Architecture}

Here, we describe the model architecture discussed in \hyperref[sec:e1]{\textbf{Sec. 4}}. Our architecture performs the following operations on an input image:

\begin{enumerate}
    \item 2-D convolution on image
    \item ReLU on output of (1)
    \item 2-D Convolution on output of (2)
    \item ReLU on output of (3)
    \item 2-D max pool on output of (4)
    \item 2-D dropout on output of (5)
    \item Fully connected layer on output of (6)
    \item ReLU on output of (7)
    \item 2-D dropout on output of (8)
    \item Fully connected layer on output of (9)
\end{enumerate}

Hence, our model architecture follows from LeNet~\cite{lecun-98}, where two convolutional layers, max pooling, ReLUs, and final fully connected layers are used.

\label{sec:a2}
\section{Additional Baselines}

In this section, we provide additional baseline results on other combinations of data sets and model architectures. Namely, we provide our MNIST~\cite{lecun-98} baselines; the low-data MNIST baselines to compare \textsc{BALD}~\cite{houlsby2011bayesian}, \textsc{VAAL}~\cite{sinha2019variational}, and adversarial DeepFool~\cite{ducoffe2018adversarial}; additional CIFAR-10 baselines using a VGG-19~\cite{simonyan2014very} model and a DLA~\cite{Yu_2018_CVPR} model; and additional SVHN baselines using a VGG-19.

\subsection{MNIST Baselines}

\begin{figure}[H]
    \centering
    \includegraphics[width=0.75\linewidth]{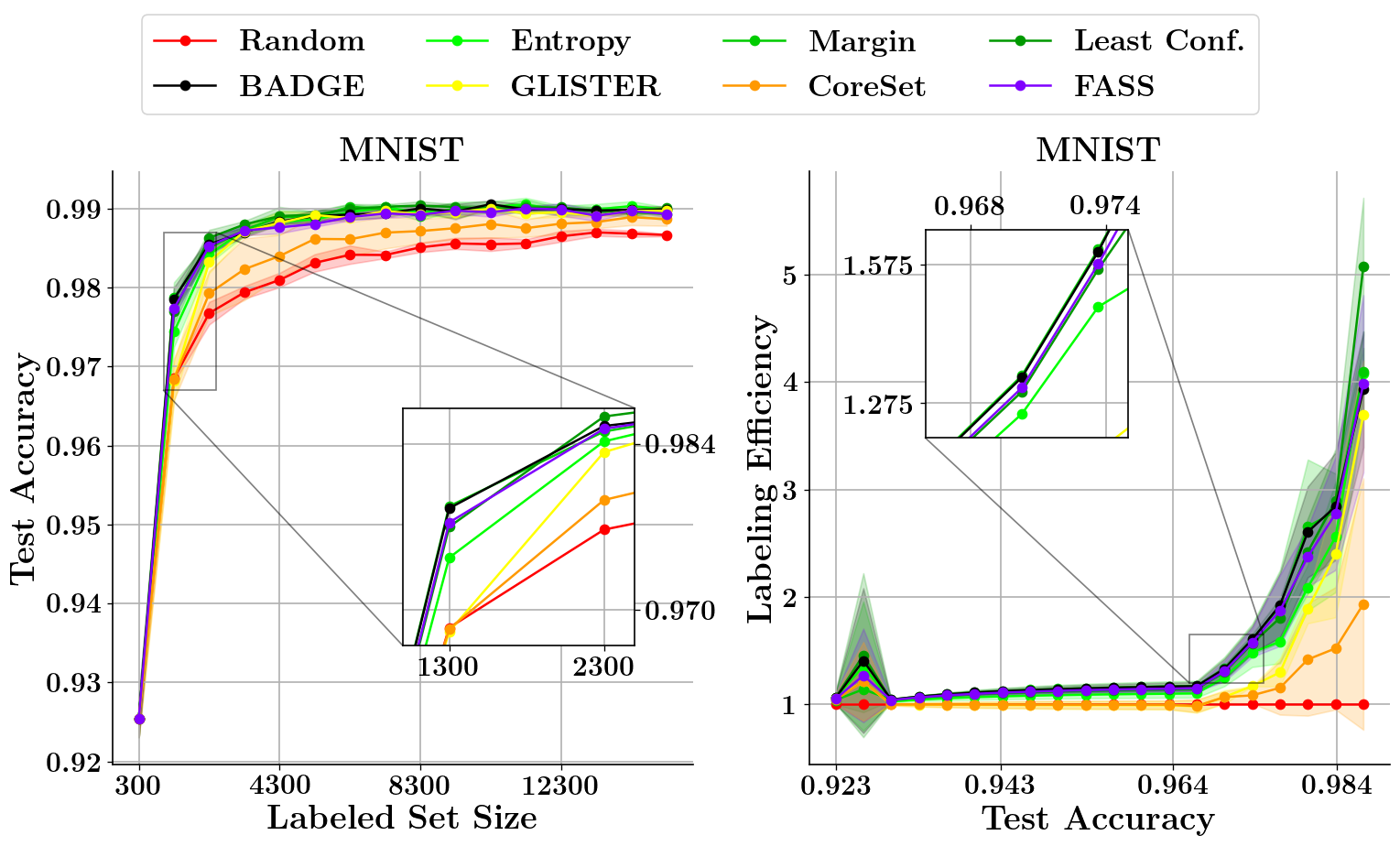}
    \caption{Baseline results with MNIST using our MNIST model architecture.}
    \label{fig:baselinemnist}
\end{figure}

The results of our MNIST baseline are given in Figure \ref{fig:baselinemnist}. Of all our baselines, we achieve the highest labeling efficiencies in MNIST. As with the other baselines, the uncertainty-based strategies are competitive with \textsc{Badge}~\cite{ash2020deep}.

\subsection{MNIST Low-Data Baselines}

\begin{wrapfigure}{r}{0.5\textwidth}
    \centering
    \includegraphics[width=0.5\textwidth]{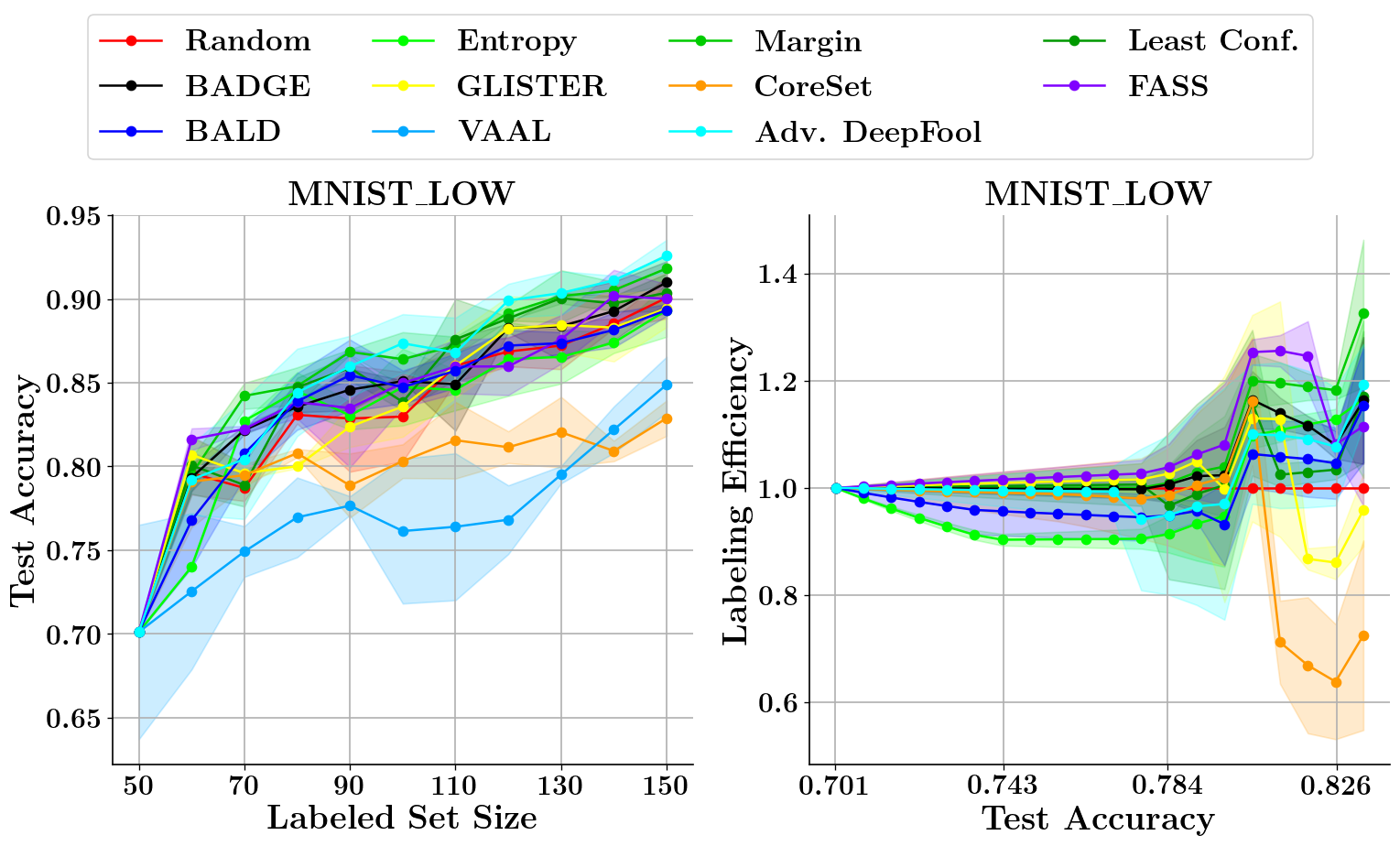}
    \caption{Baseline results with MNIST using our MNIST model architecture on very low set sizes. We see that adversarial DeepFool~\cite{ducoffe2018adversarial} performs the best out of the newly added strategies.}
    \label{fig:baselinemnistlow}
\end{wrapfigure}

\begin{wraptable}{r}{0.65\linewidth}
    \centering
    \small{
    \begin{tabular}{|l|c|}
        \hline
        \textbf{MNIST Low} & \textbf{Average Selection Time}\\
        \hline
        \textsc{Random} & 0.001 \\
        \hline
        \textsc{Entropy} & 15.482 \\
        \hline
        \textsc{Glister} & 62.685 \\
        \hline
        \textsc{Badge} & 23.277 \\
        \hline
        \textsc{FASS} & 62.415\\
        \hline
        \textsc{Least Conf} & 15.485\\
        \hline
        \textsc{Margin} & 16.495\\
        \hline
        \textsc{CoreSet} & 26.539\\
        \hline
        \textsc{BALD} & 528.519\\
        \hline
        \textsc{VAAL} & 34.218\\
        \hline
        \textsc{A. DeepFool} & 1587.458\\
        \hline
    \end{tabular}}
    \caption{Selection times for each AL strategy in each selection round (in seconds). The four strategies not compared in the main sections take much longer to select instances than the others. \textsc{VAAL} does not take that much time, but it requires approximately $4\times$ the time spent in training.}
    \label{tab:mnistlowtiming}
\end{wraptable}
To compare other recent AL strategies that did not scale well in our setting, we rerun our MNIST baseline using small budgets and low labeled set sizes. Namely, we analyze the performance of \textsc{BALD}~\cite{houlsby2011bayesian}, \textsc{VAAL}~\cite{sinha2019variational}, and adversarial DeepFool~\cite{ducoffe2018adversarial}. In this baseline experiment, we use the code provided by each work whenever applicable in our setting. The results of our baseline experiment in this setting are given in Figure \ref{fig:baselinemnistlow}. We see that the only strategy (among the ones not considered in the main paper) to perform significantly well in our baseline setting is adversarial DeepFool~\cite{ducoffe2018adversarial}.

VAAL~\cite{sinha2019variational} possibly does not do well in our setting because there is not enough time to train the VAE or the discriminator models before the task model reaches 0.99 training accuracy. Furthermore, \textsc{BALD}~\cite{houlsby2011bayesian} does not perform well against the other strategies.

Lastly, we provide the average selection times for each AL strategy for this experiment in Table \ref{tab:mnistlowtiming}. We see that three of the four strategies not considered in the main sections have astronomically higher selection times compared to the other AL strategies. While VAAL has a low selection time, it requires $4\times$ time spent in training compared to the other strategies. This is mainly because in VAAL, we train two auxilliary models -- the VAE and the discriminator -- along with the main classification model. To be concrete, take the example of CIFAR-10. The cumulative training time of AL on CIFAR-10 with a AL batch size of 3000 and with 8 rounds of AL is 8 hours for most AL algorithms. In the case of VAAL, this cumulative time will be approximately 32 hours!

To conclude, we observe that VAAL, \textsc{BALD}, and adversarial DeepFool take very long for either selection or model training. On the other hand, there is not a significant gain in terms of performance (at least in the case of MNIST with small labeled set sizes), which suggests that entropy sampling and \textsc{Badge} are probably better algorithms for AL in most cases. 

\subsection{Additional CIFAR-10 and SVHN Baselines}

\begin{figure}[h]
    \centering
    \includegraphics[width=\linewidth]{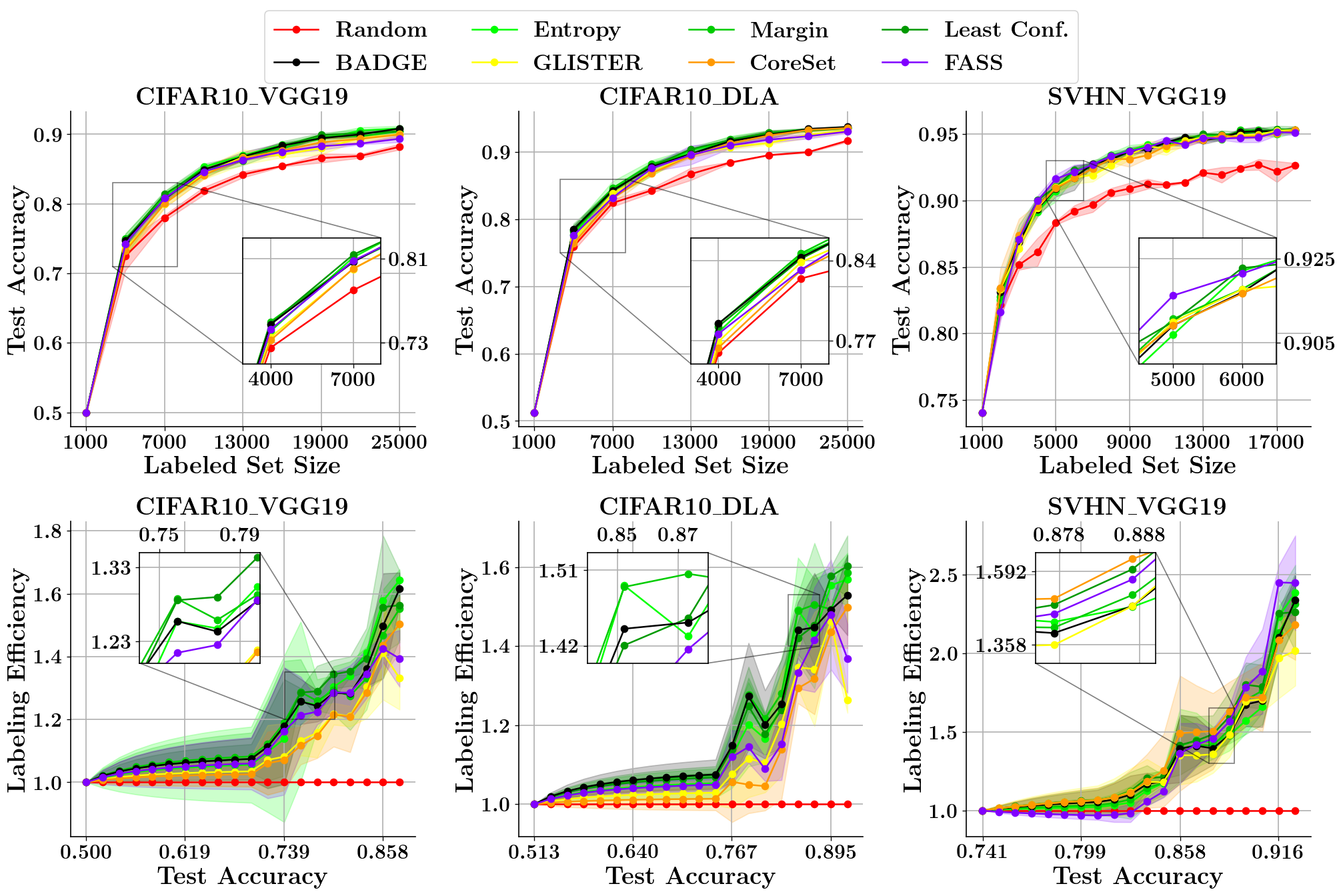}
    \caption{Additional baseline results for CIFAR-10 using VGG-19~\cite{simonyan2014very} and DLA and for SVHN using only VGG-19. Largely, the takeaways remain the same.}
    \label{fig:baselineadd}
\end{figure}

To introduce other model architectures into our baseline comparisons for robustness, we repeat the same experimental procedure for a VGG-19~\cite{simonyan2014very} model and a DLA~\cite{Yu_2018_CVPR} model for CIFAR-10~\cite{Krizhevsky09learningmultiple}. For SVHN~\cite{svhncite}, we only repeat with a VGG-19 model. The results of these experiments are given in Figure \ref{fig:baselineadd}. Largely, the takeaways remain the same: For CIFAR-10, the labeling efficiency nears $1.6\times$ (for both VGG and DLA), and the uncertainty-based AL techniques perform well against state-of-the-art techniques such as \textsc{Badge}~\cite{ash2020deep}. While the labeling efficiency is slightly reduced for SVHN, the labeling efficiency is still appreciably high (nearing $2.5\times$), and the uncertainty-based AL techniques perform well against state-of-the-art techniques such as \textsc{Badge}~\cite{ash2020deep}.

\label{sec:a3}
\section{Other Generalization Techniques}

In this section, we provide further insights on the effect of other generalization approaches. The first of these techniques is stochastic weight averaging (SWA)~\cite{izmailov2019averaging}, which computes the learned weights of a model as the average of weights periodically obtained via SGD. The second of these techniques is shake-shake (SS) regularization~\cite{gastaldi2017shakeshake}, which can be used to replace the standard sum of branches in existing residual networks with a stochastic affine combination of the branches. Both techniques have empirically shown to improve generalization performance, and the effect of each on AL performance is studied here. In this experiment, we use the same setting as the CIFAR-10 baseline experiment; however, the SWA~\cite{izmailov2019averaging} experiments use a SWA optimizer after reaching 0.95 training accuracy, and the shake-shake regularization~\cite{gastaldi2017shakeshake} experiments use a shake-shake-regularized residual network.

\begin{figure}[h]
    \centering
    \includegraphics[width=0.75\linewidth]{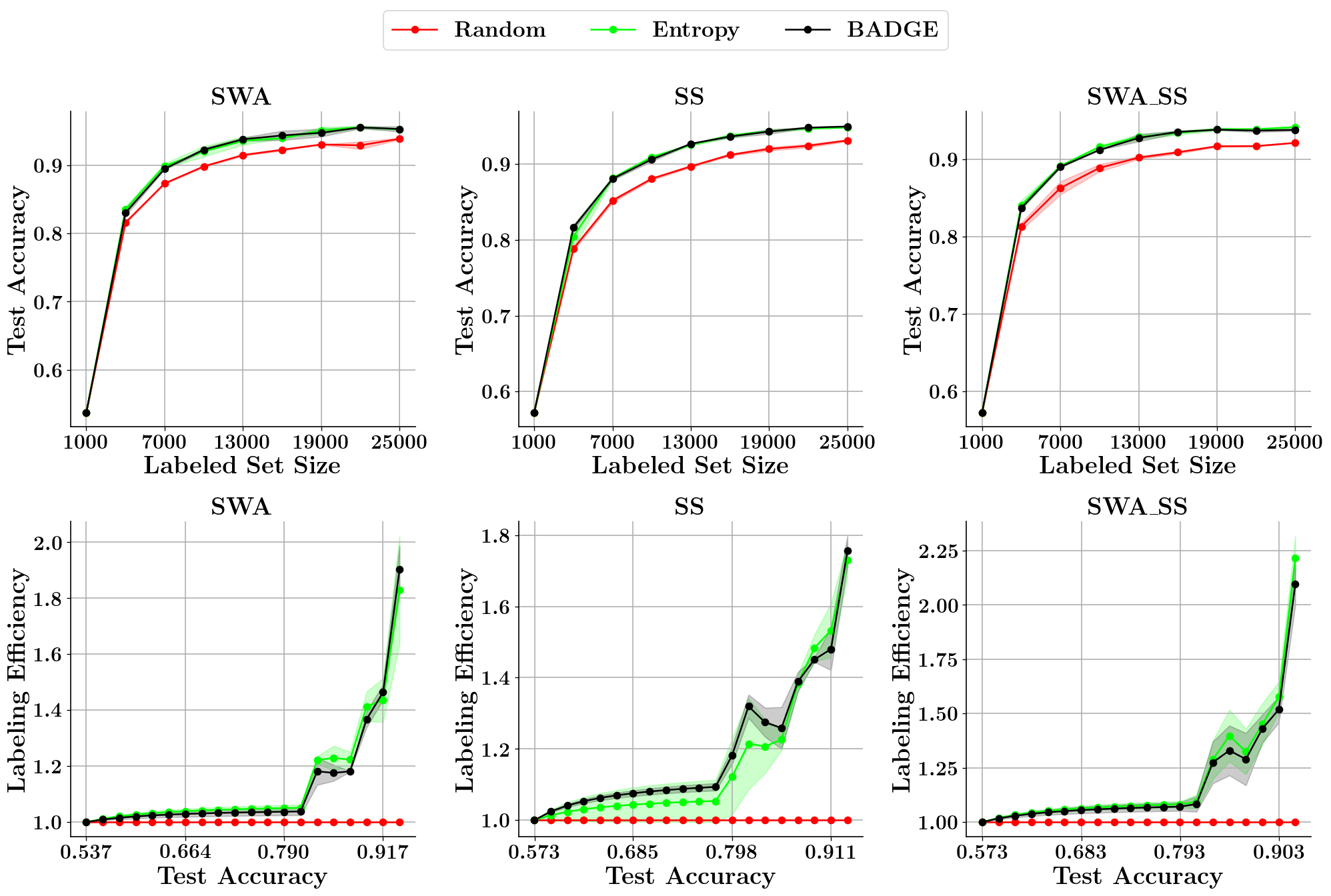}
    \caption{Effect of SWA and shake-shake (SS) regularization on our baseline settings. We see that both increase the generalization performance and do not negatively affect the labeling efficiencies.}
    \label{fig:generalization}
\end{figure}

The results of this experiment in Figure \ref{fig:generalization} indicate that both contribute to increasing the generalization performance of the model while maintaining high labeling efficiencies. Hence, we conclude that \emph{the generalization techniques studied here positively contribute to the performance of AL algorithms \& their labeling efficiency}, further enhancing the need for generalization approaches as stated in \hyperref[sec:p4]{\textbf{P4}}. The same observations (as seen in \hyperref[sec:e2]{\textbf{Sec. 4.2}}) carry forward: \textbf{a)} With the use of these generalization approaches, we see improvements in the test accuracy. \textbf{b)} We see an improvement in the labeling efficiency -- in fact, with both SWA and SS, we see a labeling efficiency of up to $2.25\times$, which is higher than the labeling efficiency we saw for the baseline experiments and also for SWA and SS applied individually. \textbf{c)} Entropy sampling and \textsc{Badge} have similar performance -- in fact, entropy sampling slightly outperforms \textsc{Badge} both in performance and labeling efficiency with SWA and SS.

\label{sec:a4}
\section{Additional Redundancy Experiment}

\begin{figure}[h]
    \centering
    \includegraphics[width=\linewidth]{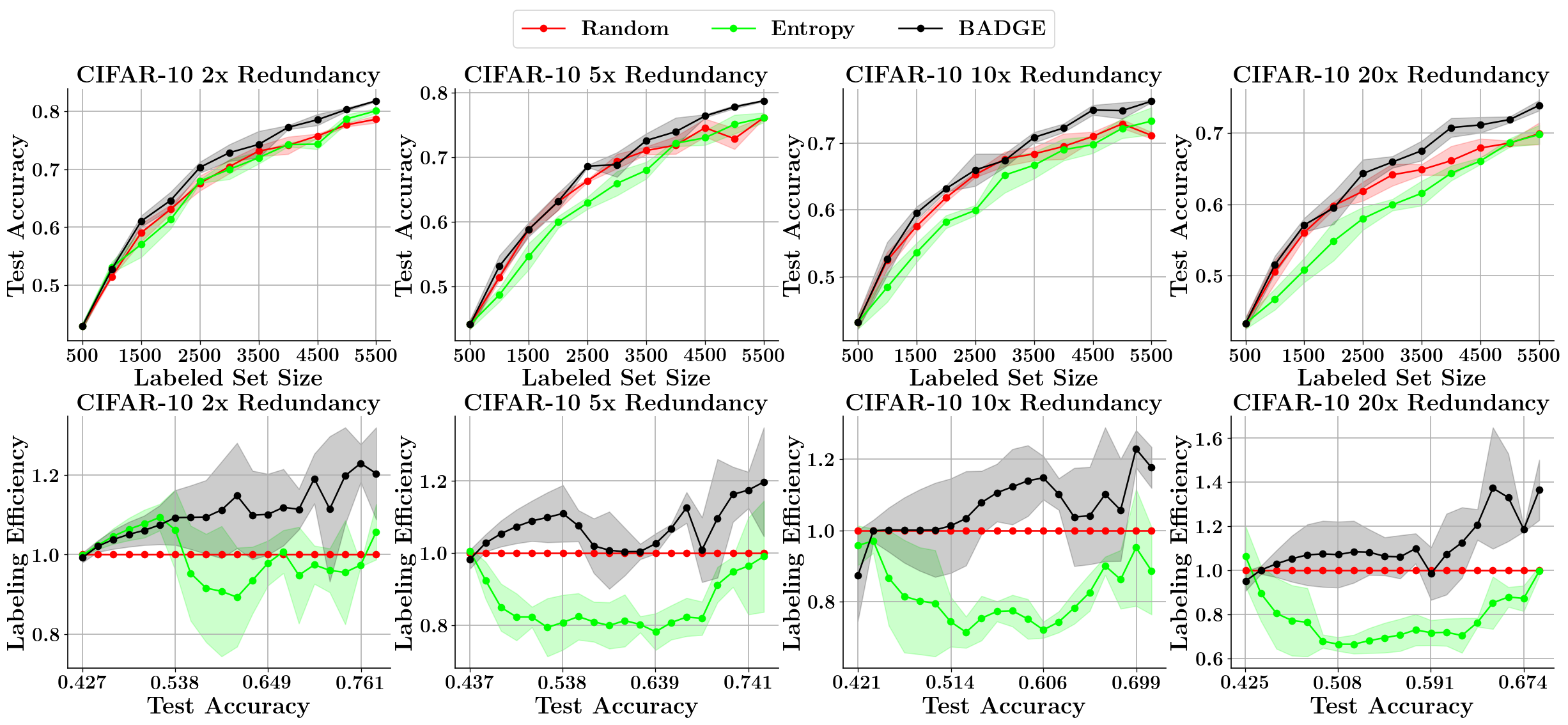}
    \caption{Performance of AL strategies on a redundant CIFAR-10 where the redundancy comes from subtle image augmentations. The takeaways are the same as in the first redundancy experiment.}
    \label{fig:appendix_red}
\end{figure}

In this section, we further explore the effect of redundancy on the performance of AL in our setting. The previous redundancy experiment explored the effect of duplicating instances from CIFAR-10~\cite{Krizhevsky09learningmultiple}; however, it is often not the case that instances are simply repeated in data sets. To study a more compelling redundancy scenario, we instead formulate redundant points by applying slight random translations and random cropping to each image. The rest of the parameters of the redundancy experiment are kept the same. Largely, we see that results remain the same: Diversity-based methods such as BADGE~\cite{ash2020deep} are robust against redundant data while uncertainty-based methods such as entropy sampling~\cite{settles.tr09} poorly handle redundant data. The results are shown in Figure~\ref{fig:appendix_red}.

\label{sec:a5}
\section{Resetting with Adam and No Augmentation}

\begin{figure}[h]
    \centering
    \includegraphics[width=0.75\linewidth,height=7cm]{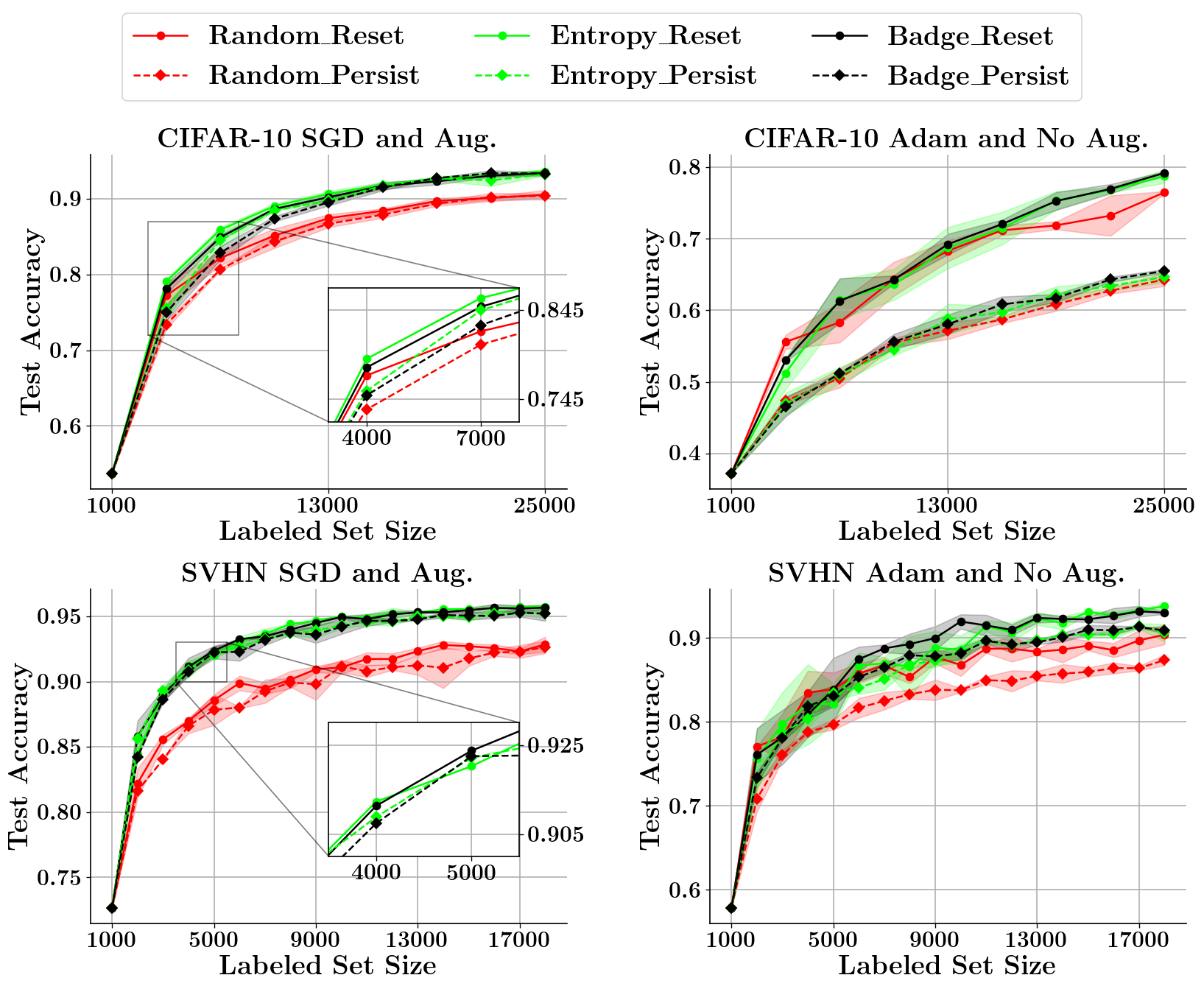}
    \caption{Test accuracies of SVHN and CIFAR-10 model resetting \textit{vs.} model updating. Here, we also provide an identical setting where Adam and no augmentation are used. We see that the results from~\cite{ash2020warmstarting} are supported by the clear performance gap in the right plots. \textbf{Note: "Persist" is synonymous with updating as the model persists across AL rounds.}}
    \label{fig:appendixreset}
\end{figure}

In our initial experiments, we compared model updating and model retraining in the scope of SGD and data augmentation. However, to corroborate past work done in~\cite{ash2020warmstarting}, we consider the impact of using Adam and no data augmentation on the performance degradation between model updating and model retraining. In this case, we repeat our updating \textit{vs.} retraining experiment where we instead use Adam and no data augmentation. From our results in Figure \ref{fig:appendixreset}, we see that the performance degradation of updating the model after each AL selection round \textit{vs.} resetting the model after each AL selection round is markedly larger, especially in CIFAR-10. Hence, we see that updating the previous model hurts generalization performance when using Adam and no augmentation, which confirms the warm-starting generalization deficiency stated in~\cite{ash2020warmstarting}. We see that in our setting, however, that the generalization gap is significantly reduced in both data sets.

\label{sec:a6}
\section{Scalability and Timing/Compute Analysis}

We briefly discussed the summary of the compute times used in our baseline experiments for \textsc{Badge} and entropy sampling. Here, we provide in Table \ref{tab:timing} the selection times and training times of that experiment. We see that the time spent in the AL loop is largely dominated by the intermediate model training and that entropy sampling takes significantly less time than \textsc{Badge} to select unlabeled instances. Specifically, the cumulative training time is around 8 hours. The cumulative selection time for \textsc{Badge} is close to one hour; in contrast, it takes only 3 minutes for entropy sampling. We also see that there is not a significant reduction if we do model updating instead of model retraining (Table \ref{tab:timing}). In particular, the cumulative training time only reduces to 7.6 hours. 

To examine other methods by which we can accelerate these training times, we conduct an experiment where most of the intermediate training rounds use adaptive DSS techniques such as \textsc{GradMatch}. To elicit the true intermediate model performance across the AL loop, we periodically use full training in a few of the AL rounds. We present in Table \ref{tab:dss} the test accuracies obtained from our \textsc{GradMatch} experiment during these full training rounds. While the test accuracies in the intermediate rounds were lower than what were reported in the CIFAR-10 baseline setting (not shown), we see that the accuracy is maintained in full training rounds. The reason for this is that there is a drop in accuracy of about 1\% with the use of an approach like \textsc{GradMatch}, but using the slightly sub-optimal model for selecting the data points does not yield a significant drop in accuracies. Hence, there is almost no difference in performance in the rounds where full training is done. Thus, we need to interleave full training to periodically gauge the performance of AL. Furthermore, we note that the intermediate training rounds complete roughly $3\times$ faster with \textsc{GradMatch} using 30\% subset sizes; hence, it makes sense to use \textsc{GradMatch} in some of the intermediate training rounds to accelerate AL. Applying these speedups to the entropy sampling times obtained in Table \ref{tab:timing} for the applicable columns, we see that the AL loop using \textsc{GradMatch} reduces to 5 hours, resulting in a $1.6\times$ speedup. We note that this speedup factor is reduced due to the full training rounds; hence, AL loops that perform full training more infrequently can appreciate larger speedups.

\begin{table}[h]
    \centering
    \small{
    \begin{tabular}{|l *{8}{|c} |}
        \hline
        \textbf{Timing} & \textbf{R1} & \textbf{R2} & \textbf{R3} & \textbf{R4} & \textbf{R5} & \textbf{R6} & \textbf{R7} & \textbf{R8} \\
        \hline
        Entropy (Reset) & 609.2 & 1397.5 & 2156.6 & 3087.2 & 3975.2 & 4940.7 & 5767.0 & 6711.6 \\
        Entropy (Update) & 498.8 & 1212.7 & 2118.0 & 3039.7 & 3823.7 & 4689.5 & 5531.0 & 6520.7 \\
        \hline
        \textsc{Badge} (Reset) & 613.1 & 1339.1 & 2120 & 3009.7 & 3942.3 & 4880.8 & 5753.3 & 6450.5 \\
        \textsc{Badge} (Update) & 417.7 & 1083.1 & 1874.6 & 2862.6 & 3691.3 & 4540.6 & 5353.2 & 6224.2\\
        \hline
        Entropy (Sel.) & 32.3 & 30.9 & 28.4 & 26.2 & 24.2 & 22.4 & 20.6 & 18.4\\
        \textsc{Badge} (Sel.) & 620.5 & 572.6 & 535.9 & 495.3 & 456.9 & 419.3 & 386.6 & 343.1\\
        \hline
    \end{tabular}}
    \caption{Time spent in training and selection (seconds) for \textsc{Badge} and entropy sampling in our CIFAR-10 baseline. The first two rows show the training times spent by entropy sampling and \textsc{Badge} when using model updating and model resetting. The third row shows the time spent by each AL strategy during their selection phases.}
    \label{tab:timing}
\end{table}

\begin{table}[h]
    \centering
    \vspace{-2ex}
    \begin{tabular}{|l*{4}{|c}|}
        \hline
        \textbf{DSS} & \textbf{1000 points} & \textbf{10000 points} & \textbf{19000 points} & \textbf{25000 points} \\
        \hline
        \textsc{Random} (GM) & $53.7 \pm 0.0$ & $85.2 \pm 0.4$ & $89.7 \pm 0.3$ & $90.5 \pm 0.5$ \\
        \textsc{Random} (F) & $53.7 \pm 0.0$ & $85.1 \pm 0.9$ & $89.7 \pm 0.4$ & $90.5 \pm 0.6$ \\
        \hline
        \textsc{Entropy} (GM) & $53.7 \pm 0.0$ & $88.5 \pm 0.1$ & $92.6 \pm 0.4$ & $93.5 \pm 0.5$ \\
        \textsc{Entropy} (F) & $53.7 \pm 0.0$ & $89.0 \pm 0.3$ & $92.6 \pm 0.1$ & $93.6 \pm 0.2$ \\
        \hline
        \textsc{Badge} (GM) & $53.7 \pm 0.0$ & $87.9 \pm 0.5$ & $92.2 \pm 0.6$ & $93.3 \pm 0.6$ \\
        \textsc{Badge} (F) & $53.7 \pm 0.0$ & $88.7 \pm 0.1$ & $92.3 \pm 0.4$ & $93.4 \pm 0.4$ \\
        \hline
    \end{tabular}
    \caption{\textsc{GradMatch} at full training rounds \textit{vs.} baseline setting at those rounds for CIFAR-10. \textsc{GradMatch} maintains comparable accuracy while being significantly faster in the other rounds not shown.}
    \vspace{-3ex}
    \label{tab:dss}
\end{table}

\label{sec:a7}
\section{Significance Tests}

The significance tests done on the results of our experiments are modeled after the penalty matrix computation done in~\cite{ash2020deep}. Namely, the significance results reported in the main sections of this paper are derived off the cells of each matrix. For clarity, the process of computing these matrices is described, and explanations are given for the claims made in the paper.

The pairwise penalty matrices (PPMs) computed in this section follow the strategy used in~\cite{ash2020deep}. In their strategy, a penalty matrix is constructed for each data set and model pair. Each cell $(i,j)$ of the matrix reflects the fraction of training rounds that AL with selection algorithm $i$ has higher test accuracy than AL with selection algorithm $j$ with statistical significance. As such, the average difference between the test accuracies of $i$ and $j$ and the standard error of that difference are computed for each training round. A two-tailed $t$-test is then performed for each training round: If $t>t_\alpha$, then $\frac{1}{N_{train}}$ is added to cell $(i,j)$. If $t<-t_\alpha$, then $\frac{1}{N_{train}}$ is added to cell $(j,i)$. Hence, the full penalty matrix gives a holistic understanding of how each selection algorithm compares against the others: A row with mostly high values signals that the associated selection algorithm performs better than the others; however, a column with mostly high values signals that the associated selection algorithm performs worse than the others.

\begin{figure}[h]
    \centering
    \begin{subfigure}[t]{0.45\textwidth}
        \includegraphics[width=\textwidth]{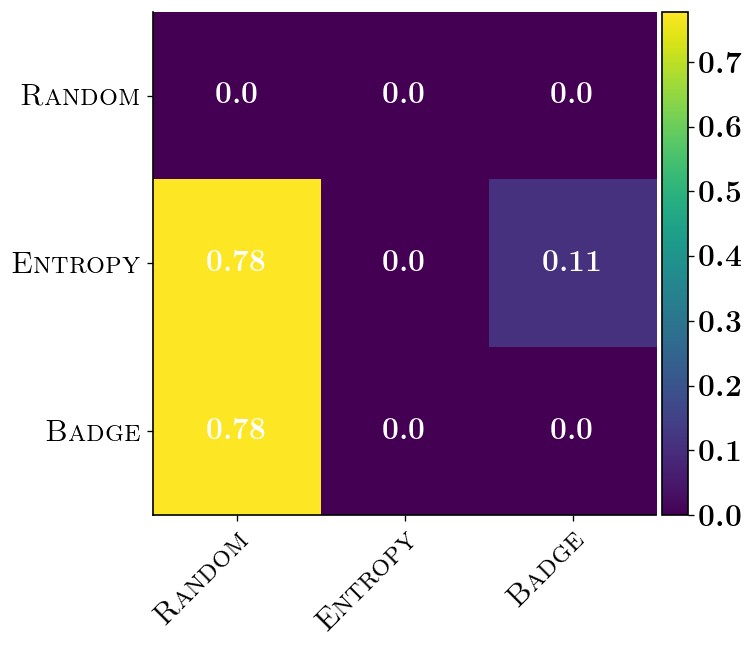}
        \vspace{-4ex}
        \caption{PPM for \hyperref[item:c1]{\textbf{Claim 1}}}
        \label{fig:ppm}
        \vspace{1ex}
    \end{subfigure}
    \begin{subfigure}[t]{0.45\textwidth}
        \includegraphics[width=\textwidth]{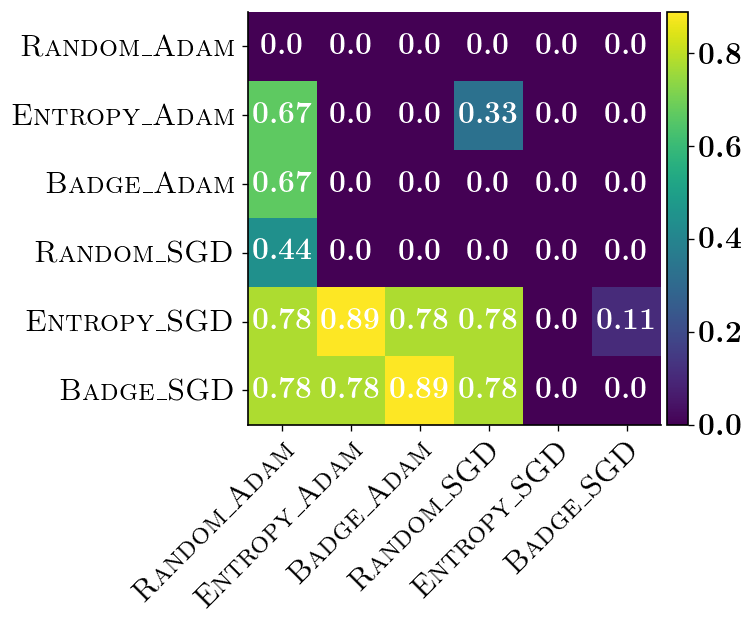}
        \vspace{-4ex}
        \caption{PPM for \hyperref[item:c2]{\textbf{Claim 2}}.}
        \label{fig:ppm2}
    \end{subfigure}
    \vspace{2ex}
    \caption{PPMs (pair-wise penalty matrices) for first two claims. The left shows that only one round of entropy sampling performs better than \textsc{Badge}. The right shows that entropy sampling with SGD performs better than entropy sampling with Adam in all but one round.}
\end{figure}

\begin{figure}[h]
    \centering
    \begin{subfigure}[t]{0.45\textwidth}
        \includegraphics[width=\textwidth]{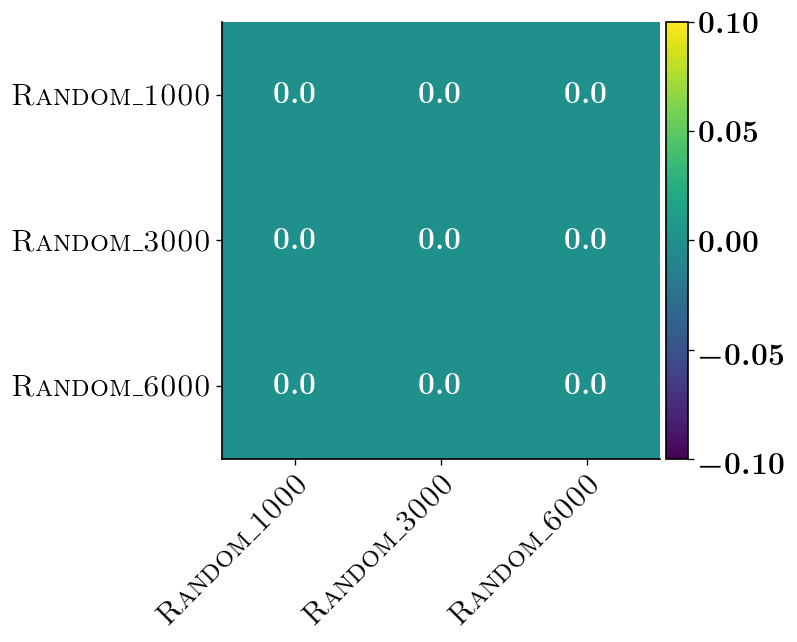}
        \includegraphics[width=\textwidth]{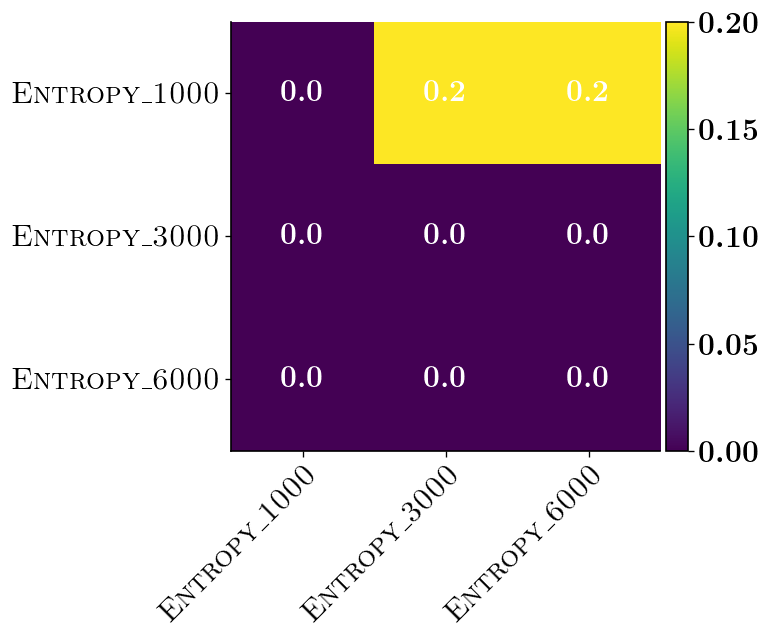}
        \includegraphics[width=\textwidth]{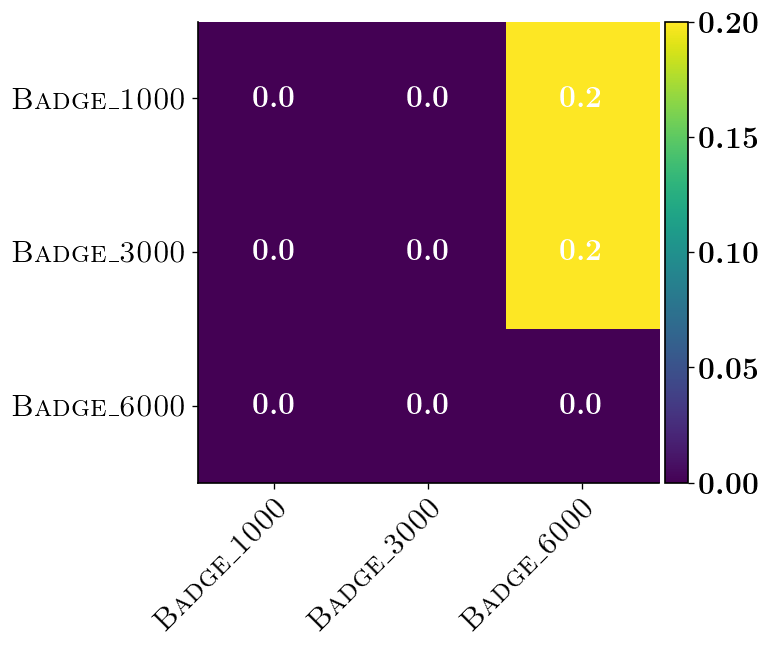}
        \caption{PPMs for \hyperref[item:c3]{\textbf{Claim 3}}.}
        \label{fig:ppm3}
    \end{subfigure}
    \begin{subfigure}[t]{0.45\textwidth}
        \includegraphics[width=\textwidth]{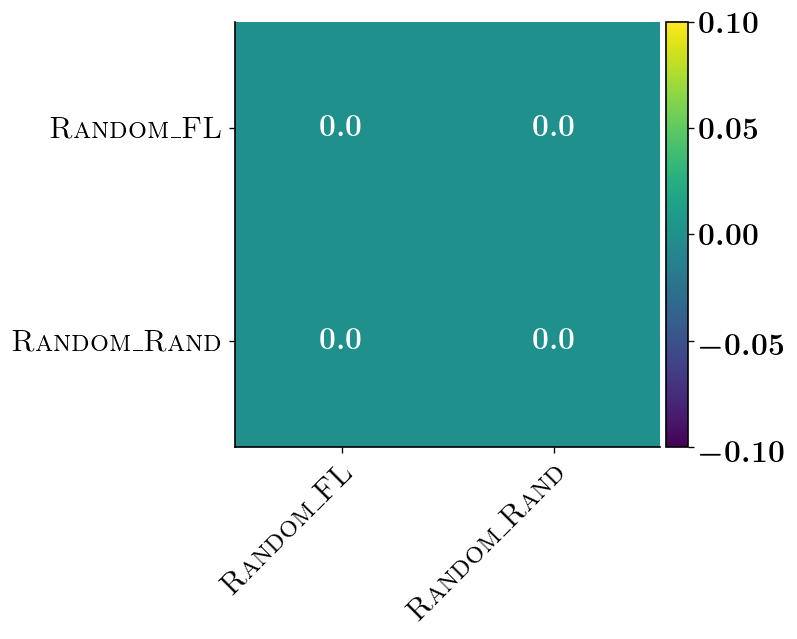}
        \includegraphics[width=\textwidth]{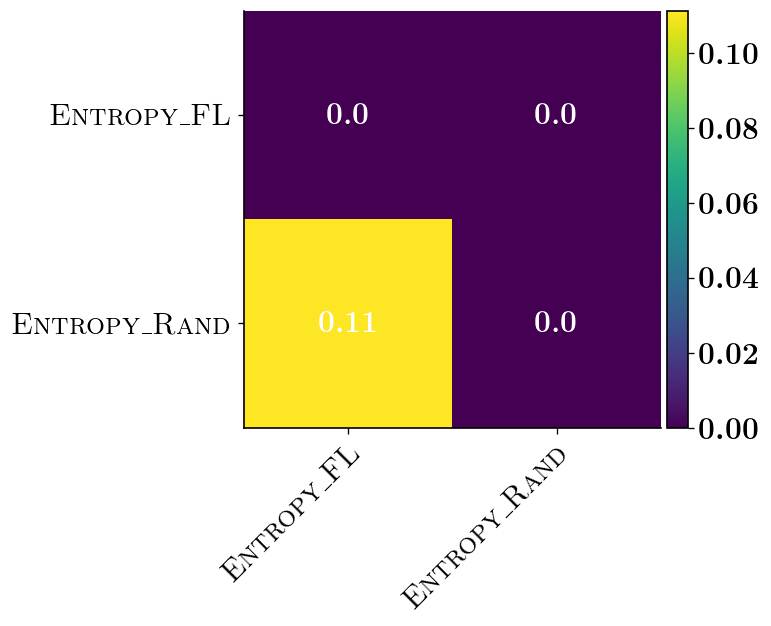}
        \includegraphics[width=\textwidth]{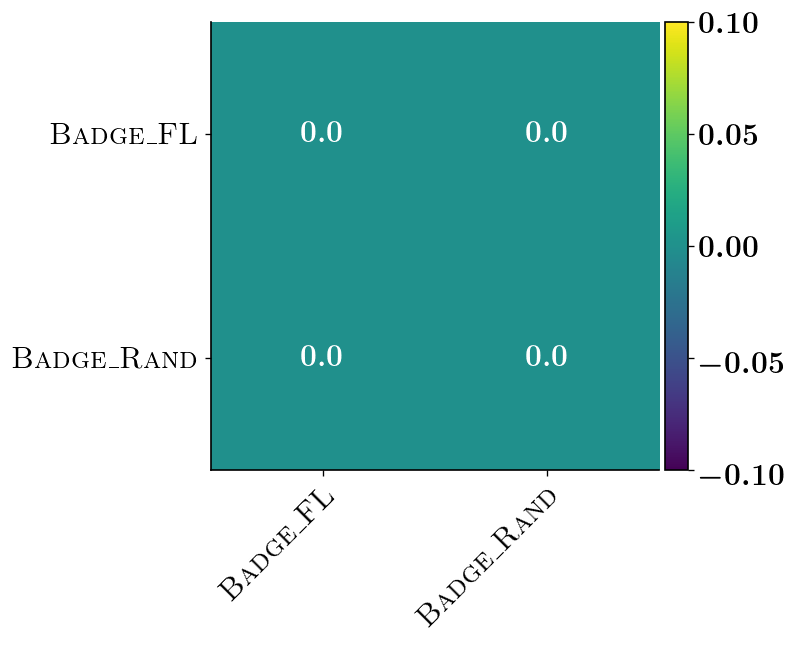}
        \caption{PPMs for \hyperref[item:c4]{\textbf{Claim 4}}.}
        \label{fig:ppm4}
    \end{subfigure}
    \label{fig:my_label}
    \vspace{1ex}
    \caption{PPMs for the third and fourth claims for random sampling, \textsc{Badge}, and \textsc{Entropy}. While it makes sense that there is no difference for random sampling, we see very little difference even for \textsc{Entropy} and \textsc{Badge} across different batch sizes (left) and initializations (right). Note: Dmin refers to dispersion-min while FL refers to facility location.}
\end{figure}

\begin{figure}[t]
    \centering
    \begin{subfigure}[t]{0.45\textwidth}
        \includegraphics[width=\textwidth]{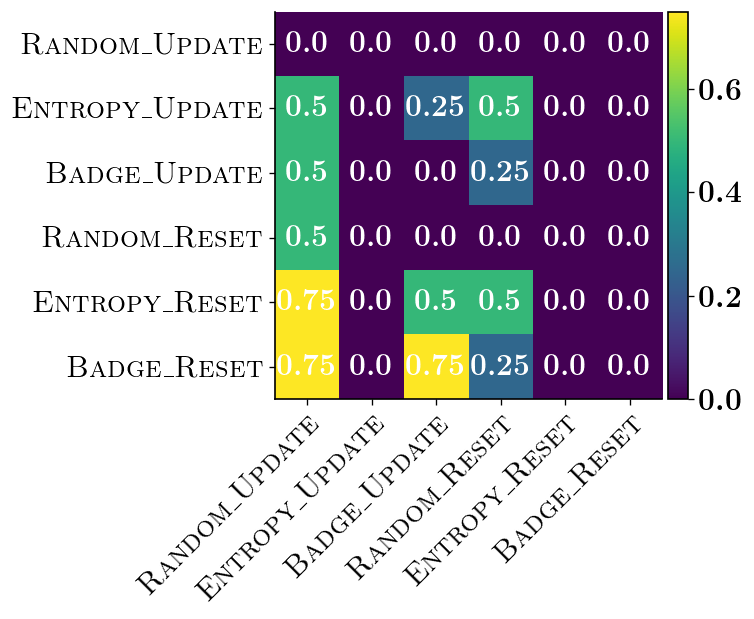}
        \caption{PPM for \hyperref[item:c1]{\textbf{Claim 5}}.}
        \label{fig:ppm5}
    \end{subfigure}
    \begin{subfigure}[t]{0.45\textwidth}
        \includegraphics[width=\textwidth]{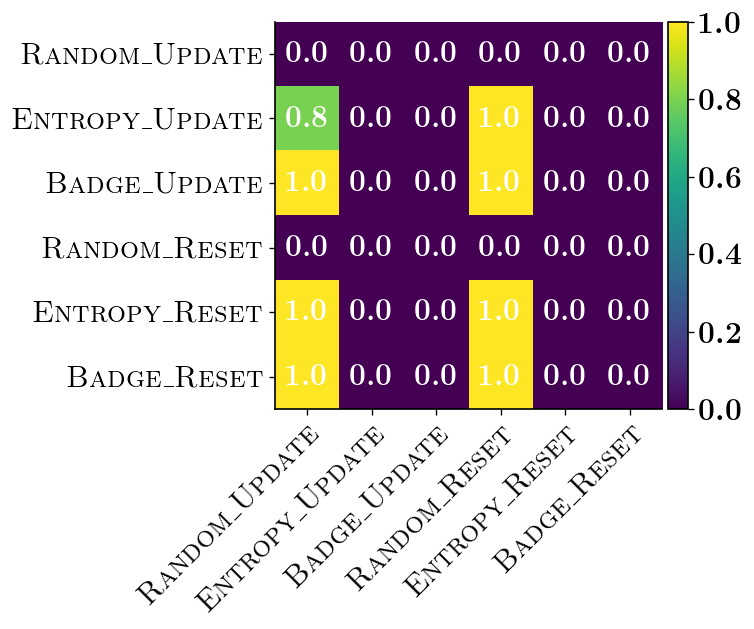}
        \caption{PPM for \hyperref[item:c1]{\textbf{Claim 5}} (2).}
    \end{subfigure}
    \vspace{2ex}
    \caption{The first PPM is calculated for the first four AL rounds; the second PPM is calculated for the remaining AL rounds. We see from the second matrix that there is no significant difference in test accuracies between updating and resetting for a given AL strategy.}
\end{figure}

In our case, we use $\alpha = 0.05$ for our $t$-tests. The relevant matrices are computed for each of the following claims, and explanations for each claim are also given:
\begin{itemize}
    \label{item:c1}
    \item \textbf{Claim 1:} Entropy sampling performs better than \textsc{Badge} in just one of the selection rounds with $\alpha = 0.05$ (\hyperref[sec:e2]{\textbf{Sec 4.2}}). In Figure \ref{fig:ppm}, cell (\textsc{Entropy},\textsc{Badge}) reads 0.11, and there were only 9 training rounds, so entropy sampling does better than \textsc{Badge} in just one round. In all other rounds, there is no statistically significant difference between the two.
    \label{item:c2}
    \item \textbf{Claim 2:} Entropy sampling with SGD scores higher than with Adam in all rounds except the initial round with $\alpha = 0.05$ (\hyperref[sec:e3]{\textbf{Sec 4.3}}). In Figure \ref{fig:ppm2}, cell (\textsc{Entropy\_SGD},\textsc{Entropy\_Adam}) reads 0.89, and there were only 9 training rounds, so entropy sampling with SGD does better than entropy sampling with Adam in all but one round. Since the initial accuracy is the same across each configuration, it must be the initial round.
    \label{item:c3}
    \item \textbf{Claim 3:} No two configurations of the AL batch size had accuracies significantly different with $\alpha = 0.05$ for more than 20\% of the selection rounds (\hyperref[sec:e6]{\textbf{Sec 4.6}}). In Figure \ref{fig:ppm3}, all cells are a maximum of 0.2, which confirms that the 5 comparable rounds (1000, 7000, 13000, 19000, 25000) only had significant differences in 20\% of the rounds.
    \label{item:c4}
    \item \textbf{Claim 4:} Entropy sampling with a random initialization versus entropy sampling with a facility location initialization is the only comparison to show a significant difference in test accuracy in a selection round with $\alpha = 0.05$ (\hyperref[sec:e6]{\textbf{Sec 4.6}}). As shown in Figure \ref{fig:ppm4}, the only cell that is not zero is (\textsc{Entropy\_Rand},\textsc{Entropy\_FL}); hence, the claim follows.
    \label{item:c5}
    \item \textbf{Claim 5:} There is not a statistically significance difference in test accuracy with $\alpha = 0.05$ between resetting or updating for any given strategy in the later rounds (\hyperref[sec:e7]{\textbf{Sec 4.7}}). In Figure \ref{fig:ppm5}, the first matrix shown is computed for the first four training rounds while the second matrix shown is computed for the last five training rounds. As shown, each (\textsc{*\_Update}, \textsc{*}) cell in the latter matrix is 0, and so is each (\textsc{*}, \textsc{*\_Update}) cell. Hence, the claim follows.
\end{itemize}

\label{sec:a8}
\section{Societal Impacts and Limitations}

It is important to consider the limitations of this work and the potential societal impacts that this work has. Perhaps the largest limitation of this work is its reliance on data augmentation. Indeed, not all tasks in AL afford the use of data augmentation. In our case, we studied the effect of augmenting image data for classification tasks; however, many other AL domains such as text-based tasks do not have nearly the same capability of data augmentation. Hence, many of the observations in this paper only apply to domains that readily accept data augmentation, which typically applies to image data. Another limitation is that the results of this paper are prefaced on academic data sets and modifications thereof; applicability to other image classification data sets may not always follow the trends mentioned in this paper. However, we highly suspect that the trends mentioned in this paper apply to most image classification tasks where data augmentation can be done.

Although it is not a main talking point in this work, effectively evaluating AL algorithms does have societal impacts. Namely, not effectively evaluating AL algorithms can present misguided results about which AL algorithm performs well, which can have adverse effects on everything benefiting from AL tasks. As an example, since many AL tasks now operate on images of people, it is imperative for AL algorithms to achieve state-of-the-art generalization performance to prevent accidental discrimination. If an AL algorithm is not evaluated effectively using the generalization techniques discussed in this paper, then its actual performance in industry might suffer, resulting in worse generalization performance compared to what could have been achieved using other AL methods. By effectively evaluating AL algorithms in our setting, we can ensure that the results on each AL algorithm are faithful to all the practices done in industry to achieve good generalization performance.

Finally, as discussed in the conclusion of the main paper, the suggestions for better generalization approaches (\textit{e.g.}, the use of data augmentation and SGD in the training loop or even more complicated SWA/SS approaches) come with a significantly higher cost of compute. To achieve better generalization, we essentially need to train the models for more epochs, which results in higher experimental turn-around times, higher energy consumption, and higher CO2 emissions. We hope that many of the suggestions here like the use of data subset selection and specifically \textsc{GradMatch}~\cite{killamsetty2021gradmatch} can help in reducing the turn-around times, energy consumption, and CO2 emissions. 


\end{document}